\documentclass[journal]{IEEEtran}
\usepackage{cite}
\usepackage{graphicx}
\usepackage{amsmath}
\usepackage{amssymb}
\usepackage{subfigure}
\usepackage{stfloats}
\usepackage{algorithm}
\usepackage{algorithmicx}
\usepackage{algpseudocode}
\usepackage{url}
\usepackage{tabularx}
\usepackage{array}
\usepackage{multirow}
\usepackage{color}
\usepackage{booktabs}
\usepackage{enumerate}
\usepackage{breqn}
\usepackage{indentfirst}
\usepackage{bm}
\usepackage{mathtools}

\begin{document}
	\title{E$ \mathbf{^3} $MoP: Efficient Motion Planning Based on Heuristic-Guided Motion Primitives Pruning and Path Optimization With Sparse-Banded Structure}
	
	\author{\IEEEauthorblockN{Jian Wen, Xuebo Zhang,~\IEEEmembership{Senior Member,~IEEE,} Haiming Gao, Jing Yuan,~\IEEEmembership{Member,~IEEE,} \\and Yongchun Fang,~\IEEEmembership{Senior Member,~IEEE}}
	\thanks{This work is supported in part by National Key Research and Development Project under Grant 2018YFB1307503, in part by Tianjin Science Fund for Distinguished Young Scholars under Grant 19JCJQJC62100, in part by Tianjin Natural Science Foundation under Grant 19JCYBJC18500, and in part by the Fundamental Research Funds for the Central Universities. \emph{(Corresponding author: Xuebo Zhang.)}}
	\thanks{Jian Wen, Xuebo Zhang, Jing Yuan, and Yongchun Fang are with the Institute of Robotics and Automatic Information System, College of Artificial Intelligence, Nankai University, Tianjin 300350, China, and also with the Tianjin Key Laboratory of Intelligent Robotics, Nankai University, Tianjin 300350, China (e-mail: zhangxuebo@nankai.edu.cn; wenjian@mail.nankai.edu.cn).}
	\thanks{Haiming Gao is with Zhejiang Lab, Hangzhou 311121, China.}
	}

\maketitle

\begin{abstract}
	To solve the autonomous navigation problem in complex environments, an efficient motion planning approach is newly presented in this paper. Considering the challenges from large-scale, partially unknown complex environments, a three-layer motion planning framework is elaborately designed, including global path planning, local path optimization, and time-optimal velocity planning. Compared with existing approaches, the novelty of this work is twofold: 1) a novel heuristic-guided pruning strategy of motion primitives is proposed and fully integrated into the state lattice-based global path planner to further improve the computational efficiency of graph search, and 2) a new soft-constrained local path optimization approach is proposed, wherein the sparse-banded system structure of the underlying optimization problem is fully exploited to efficiently solve the problem. We validate the safety, smoothness, flexibility, and efficiency of our approach in various complex simulation scenarios and challenging real-world tasks. It is shown that the computational efficiency is improved by 66.21\% in the global planning stage and the motion efficiency of the robot is improved by 22.87\% compared with the recent quintic B\'{e}zier curve-based state space sampling approach. We name the proposed motion planning framework E$ \mathbf{^3} $MoP, where the number 3 not only means our approach is a three-layer framework but also means the proposed approach is efficient in three stages.
\end{abstract}

\def\abstractname{Note to Practitioners}
\begin{abstract}
	This paper is motivated by the challenges of motion planning problems of mobile robots. A three-layer motion planning framework is proposed by combining global path planning, local path optimization, and time-optimal velocity planning. For mobile robot navigation applications in semi-structured environments, optimization-based local planners are recommended. Extensive simulation and experimental results show the effectiveness of the proposed motion planning framework. However, due to the non-convexity of the path optimization formulation, the proposed local planner may get stuck in local optima. In future research, we will concentrate on extending the proposed local path optimization approach with the theory of homology classes to maintain several homotopically distinct local paths and seek global optima.
\end{abstract}


\IEEEpeerreviewmaketitle

\section{Introduction}
\IEEEPARstart{M}{OBILE} robots have been widely applied in various scenarios, such as family service, logistics, search-and-rescue, and so on. Motion planning is one of the key technologies for mobile robots to achieve full autonomy in these scenarios \cite{latombe1991robot, lavalle2006planning, choset2005principles}, which has been comprehensively investigated in the literature \cite{gonzalez2015review}. However, it is still challenging to design a motion planning approach that can ensure both safety, smoothness, flexibility, and efficiency in large-scale, unknown, or partially unknown complex environments \cite{alterovitz2016robot}.

For computational efficiency considerations, the commonly adopted motion planning framework is organized in a two-layer architecture, namely, global path planning and local planning \cite{lunenburg2016motion}. The global path planner is employed to provide a rough route from the current robot pose to the goal one \cite{liu2015robotic,chi2018risk,wang2020eb, wang2020neural}, followed by a local planner supposed to generate safe and smooth motions according to real-time sensor data \cite{zhou2020trajectory,li2015real, zhou2020autonomous}. In \cite{zhang2018multilevel}, Zhang \emph{et al.} present a multilevel human-like motion planning framework, wherein global-level path planning, sensor-level path planning, and action-level trajectory planning corresponding to the functions of the human brain, eyes, and legs are designed respectively. In the work \cite{marder2010office}, Marder-Eppstein \emph{et al.} propose a robust navigation system for mobile robots in office environments by combining a global planner with Dijkstra's algorithm \cite{dijkstra1959note} and a local planner with the well-known dynamic window approach (DWA) \cite{fox1997dynamic}. In \cite{wang2017autonomous}, Wang \emph{et al.} design a motion planning framework for autonomous mobile robot navigation in uneven and unstructured indoor environments, in which an improved rapidly-exploring random tree (RRT)-based global planner and the elastic bands (EB) local planner \cite{quinlan1993elastic} are employed.

In this work, an efficient motion planning framework is proposed. Different from the classical two-layer motion planning framework, local motion planning is decoupled into local path optimization and time-optimal velocity planning, and the two-layer motion planning framework is accordingly extended to a three-layer framework. Simulation and experimental results demonstrate that the proposed motion planning framework achieves better computational efficiency in both global path planning and local path optimization and better motion efficiency in velocity planning. Therefore, we name the proposed motion planning framework E\ensuremath{^3}MoP. Here, the number 3 not only means our approach is a three-layer framework but also means the proposed approach is efficient in three stages.

\subsection{Global Planning}
Sampling-based planning and grid-based planning are two popular global path planning approaches. Typical sampling-based planning algorithms include probabilistic roadmap (PRM) \cite{kavraki1996probabilistic} and RRT \cite{lavalle2001randomized}. These algorithms represent the configuration space (C-space) with a roadmap of randomly sampled configurations, which has considerable advantages in computational efficiency, especially in high-dimensional planning problems. However, sampling-based planning is limited by completeness and optimality, even some excellent variants such as RRT* can only guarantee asymptotic optimality \cite{karaman2011sampling}. In this work, we focus on grid-based path planning approaches.
 
Grid-based planning overlays a grid on C-space and assumes each configuration is identified with a grid-cell center. Then search algorithms such as Dijkstra's algorithm \cite{dijkstra1959note} and A* \cite{hart1968formal} are used to find a path from the start to the goal, which can be complete and optimal in the discrete search space. However, original grid-based planning algorithms tend to produce paths that are non-smooth and do not generally satisfy the kinodynamic constraints of the robot. In \cite{pivtoraiko2005generating}, Pivtoraiko and Kelly propose the state lattice approach for graph construction, where the connectivity between two configurations in the graph is built from motion primitives which fulfill the kinodynamic constraints of the robot. In contrast to 4-connected or 8-connected grid-based representations, the feasibility of connections in the lattice guarantees that any path found using a lattice will also be feasible. This makes it very suitable to planning for non-holonomic robotic systems. Furthermore, to improve the computational efficiency of graph search, informative heuristics are employed to guide the search towards the most promising areas of the search space \cite{hart1968formal}. The Euclidean distance function is perhaps the most well-known heuristic. However, it has no knowledge of environmental information and is thus a poor heuristic when applied in an environment with dense obstacles since it seriously underestimates the actual path costs \cite{knepper2006high}. In \cite{likhachev2009planning}, Likhachev and Ferguson propose a 2-D grid-based heuristic $ h_{2D} $, which is derived by Dijkstra's algorithm incorporating the up-to-date environmental information. $ h_{2D} $ computes the costs of shortest paths from the goal cell to other cells in the 2-D grid and captures the geometry of obstacles in the environment. However, in \cite{likhachev2009planning} $ h_{2D} $ is only employed to estimate the cost of the shortest path from a given search state to the goal state, while the information about the graph search direction behind the heuristic has not been fully exploited.

\subsection{Local Planning}
Sampling-based local planning and optimization-based local planning are two popular local planning approaches in the field of mobile robots. A sampling-based local planner usually generates a set of candidate trajectories/paths first and then selects the best one which minimizes an elaborately designed evaluation function. The evaluation function usually includes a variety of factors, for instance, the distance to obstacles and the deviation from the global path. Some sampling-based local planners work in control space, such as the popular DWA \cite{fox1997dynamic}. Compared with sampling in control space, sampling in state space is superior in terms of sampling efficiency and robustness to initial conditions \cite{howard2014model}. In the work \cite{howard2008state}, Howard \emph{et al.} propose a state space sampling approach with a vehicle model-based trajectory generation approach \cite{howard2007optimal}, which has been successfully applied in the DARPA Urban Challenge \cite{ferguson2008motion}. However, the environmental constraints are not taken into account during trajectory generation in \cite{howard2008state}, and considerable time is wasted in generating infeasible trajectories \cite{chen2014quartic}. In \cite{zhang2018multilevel}, Zhang \emph{et al.} propose a state space sampling-based local planner by generating quintic B\'{e}zier curve-based paths with different initial curvatures offline. These local paths are saved in a lookup table and retrieved in the local planning stage according to the curvature condition, thus a significant amount of time is saved for online path generation. However, the endpoints of the offline generated paths are fixed, somehow limiting the flexibility of local planning. And in an environment with dense obstacles, sampling-based local planners may even fail to find a solution.

Optimization-based local planning formulates the local planning problem as a non-linear optimization problem, which takes the global path in the local window as input and deforms the local path to make the optimization problem converge. In \cite{dolgov2010path}, Dolgov \emph{et al.} propose a conjugate gradient (CG)-based path optimization approach for autonomous vehicle-free space planning, wherein the smoothness and safety of the path and the curvature constraint of the vehicle are both considered. In the work \cite{rosmann2012trajectory,rosmann2013efficient}, R{\"o}smann \emph{et al.} propose the well-known timed elastic bands (TEB) local planner. Different from the classical EB \cite{quinlan1993elastic}, TEB explicitly considers the temporal aspects of motions, thus the initial path generated by the global path planner can be optimized with respect to minimizing the trajectory execution time, and kinodynamic constraints of robots such as maximum velocities and accelerations can be incorporated into the optimization objective as soft constraints. Inspired by the back-end of simulations localization and mapping (SLAM), TEB formulates the optimization problem in the hyper-graph framework and employs the graph optimization solver g2o \cite{kummerle2011g} to solve the problem. However, the essential \emph{banded} system structure behind the optimization problem has not been fully discussed and exploited in TEB. Furthermore, although TEB considers the velocity and acceleration constraints, it can not guarantee that these kinodynamic constraints are strictly satisfied in the soft constraint framework. In addition, too many constraints may lead to mutual compromises. For example, keeping a minimum distance to obstacles may be conflicting with acquiring a minimum-time trajectory. Therefore, the final optimized trajectory may not achieve good results in both motion efficiency and safety.

\subsection{Contributions}
Motivated by the challenges of motion planning problems of mobile robots and the aforementioned limitations of existing approaches, a three-layer motion planning framework called E\ensuremath{^3}MoP is proposed in this paper, including global path planning, local path optimization, and time-optimal velocity planning. Specifically, we propose new approaches in the module of global planning and local path optimization part:
\subsubsection{Global Planning}
The A* search algorithm combined with motion primitives is adopted in the stage of global path planning. Inspired by the environment-constrained heuristic $ h_{2D} $ presented in \cite{likhachev2009planning}, a novel heuristic-guided pruning strategy of motion primitives is proposed to further improve the computational efficiency of graph search. Given a set of motion primitives as the action set, the branching factor, i.e., the average number of successors per state, is fixed. Specifically, all the motion primitives are involved during every expanding process in the A* search. Different from using a complete set of motion primitives without pruning, in this work the search direction information behind $ h_{2D} $ is exploited to provide a one-step forward prediction for the A* search, and then motion primitives pruning is conducted so that only part of the motion primitives are involved in each expanding process. Therefore, the branching factor of graph search is decreased, and the computational efficiency is significantly improved.

\subsubsection{Local Planning}
Local motion planning problems are addressed by a path/velocity decoupled planning framework. A soft-constrained multi-objective local path optimization approach is newly proposed, wherein the constraints including safety and smoothness are both considered. Furthermore, we notice that each sub-objective only depends on a few local consecutive variables, thus the partial derivatives of the irrelevant variables in the Jacobian are all zero and the Hessian matrix of the whole optimization problem is \emph{sparse-banded}. Based on this property, the optimization problem is efficiently solved through the Levenberg–Marquardt (LM) algorithm combined with the forward elimination and back substitution algorithm, and high real-time optimization performance can be achieved. After path optimization, cubic spline interpolation is employed to further smooth the local path. Finally, a numerical integration-based velocity planning algorithm described in \cite{zhang2018multilevel} is utilized to generate a feasible linear velocity profile along the smoothed local path under the velocity and acceleration constraints.

To summarize, the main contributions of this work are as follows:
\begin{enumerate}[\hspace{1em}1)]
	\item A novel heuristic-guided pruning strategy of motion primitives is proposed, which is fully integrated into the state lattice-based global path planner to further improve the computational efficiency of graph search.
	\item A new soft-constrained local path optimization approach is proposed. The sparse-banded system structure of the underlying optimization problem is fully exploited to efficiently solve the problem, which converges quickly to generate a safe and smooth local path. The smoothness of the local path benefits the subsequent time-optimal velocity planning.
	\item Autonomous navigation is realized in large-scale, partially unknown complex environments. Extensive simulation and experimental evaluations are presented to verify the safety, smoothness, flexibility, and efficiency of the proposed approach. 
\end{enumerate}

The remainder of the paper is organized as follows. Section II introduces the system framework of autonomous navigation. The pruning strategy of motion primitives and the local path optimization approach are detailed in Sections III and IV respectively. Section V provides some implementation details. The results of simulations and experiments are presented and discussed in Sections VI and VII respectively. Finally, this paper is concluded in Section VIII. 

\section{System framework}
\begin{figure}[t]
	\centering
	\includegraphics[scale=0.3]{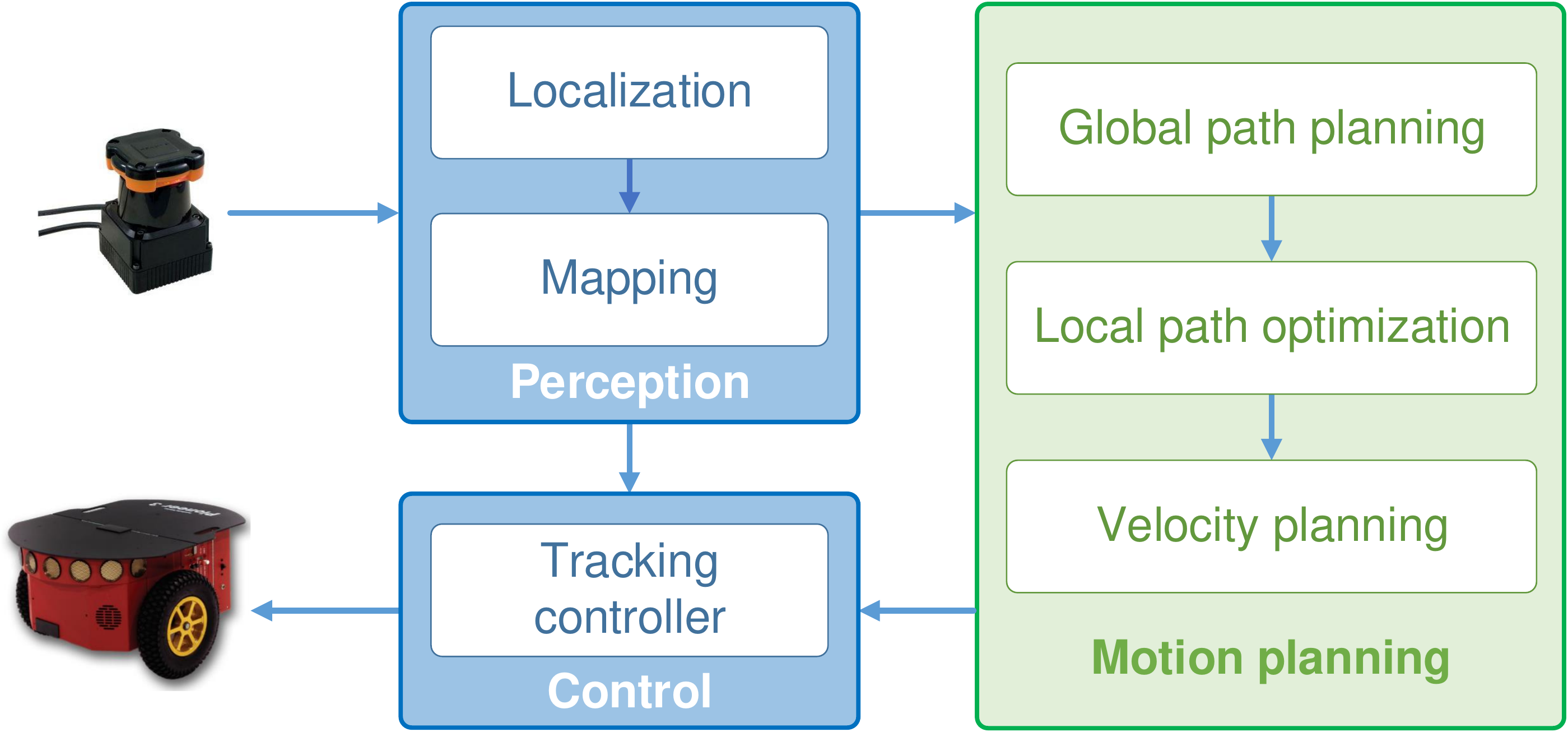}
	\caption{Software system architecture designed for autonomous navigation.}
	\label{fig:systemframework}
\end{figure}

In this work, we follow the ``see-think-act'' pipeline and design an autonomous navigation software system consisting of three primary modules, namely, perception, motion planning, and control, as illustrated in Fig. \ref{fig:systemframework}. In particular, the motion planning module of E\ensuremath{^3}MoP is detailed in Fig. \ref{fig:motionplanning}. All the modules run in parallel and communicate with each other via messages based on the Robot Operating System (ROS) \cite{quigley2009ros}. Next, we will briefly introduce some details of localization, mapping, and motion planning.

\subsection{Localization}
Perception is a fundamental module for autonomous navigation \cite{sun2020plane,cheng2020improving}. In this work, a feature-based localization approach is employed. When a new laser scan is received, an efficient and robust 2-D line segment extraction algorithm described in \cite{gao2018line} is utilized to extract line segment features. Then, the endpoints of these line segment features are matched with a prior feature map by the Random Sample Consensus (RANSAC) algorithm \cite{fischler1981random}. If the number of the matched pairs is greater than a preset threshold, the robot pose is obtained by computing the transformation between the matched point features. Otherwise, dead rocking is temporarily employed. \emph{It should be noted that the proposed navigation system is highly modular and the localization module can be replaced by other alternatives.}

\subsection{Mapping}
In this work, the environment is represented in the form of the Euclidean Distance Grid (EDG), wherein each cell stores the distance to the closest obstacle in the grid. Cells occupied by obstacles have a zero value. Such representation is convenient and efficient to evaluate whether a configuration is in collision with obstacles or not. For example, a collision-free judgment can be obtained if the minimum value of the cells covered by the robot is larger than the circumscribed radius of the robot footprint. With the help of EDG, considerable computing time for directly employing the geometric footprint of the robot to collision detection can be saved.

\begin{figure}[t]
	\centering
	\includegraphics[scale=0.33]{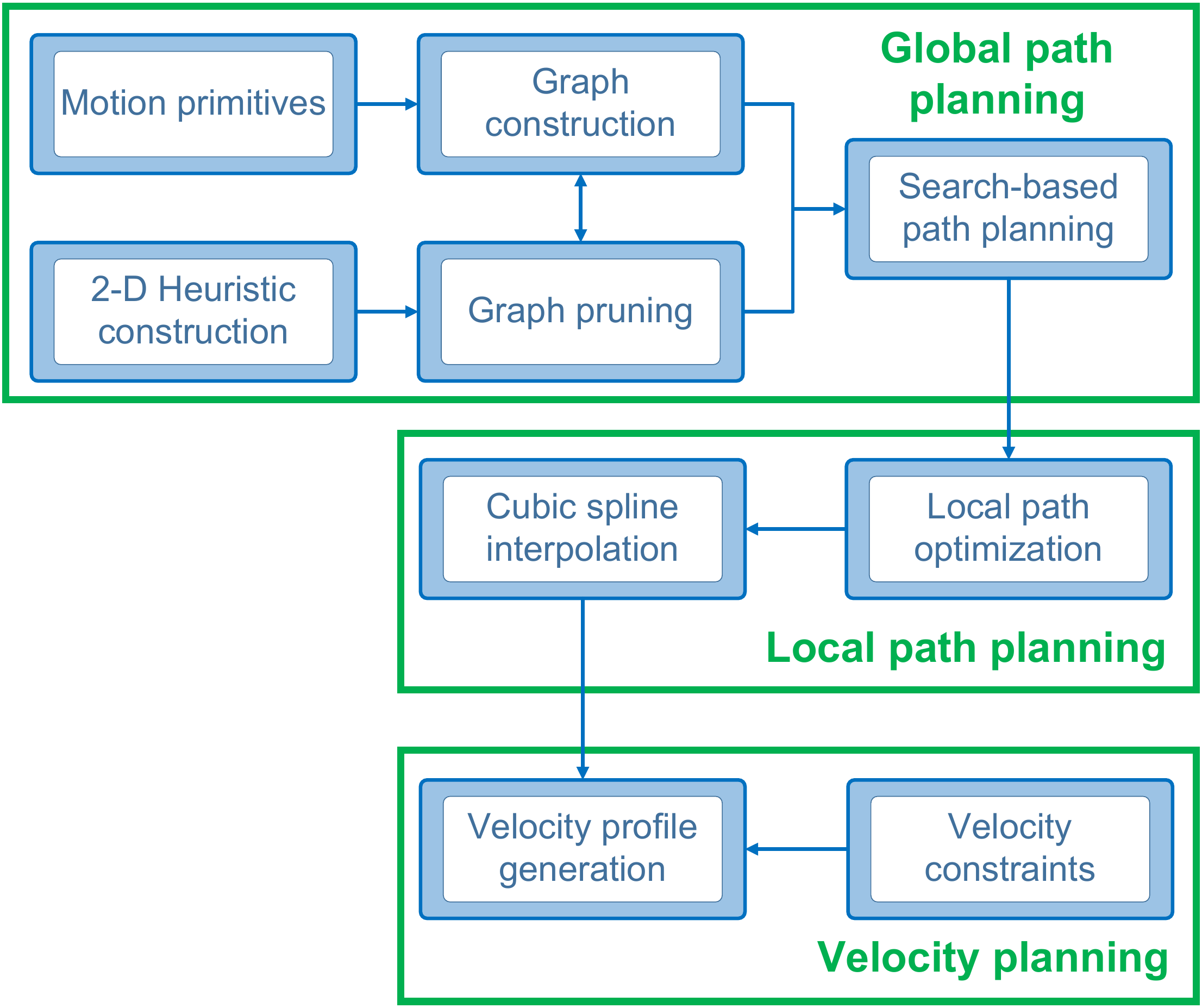}
	\caption{Flow chart of the three-layer motion planning framework E\ensuremath{^3}MoP.}
	\label{fig:motionplanning}
\end{figure}

\subsection{Motion Planning with E\ensuremath{^3}MoP}
Motion planning, which plays an essential role in generating safe, smooth, efficient, and flexible motions for mobile robots, is the main focus of this work. Considering the challenges from large-scale, partially unknown complex environments, an efficient motion planning approach called E\ensuremath{^3}MoP is newly proposed. As illustrated in Fig. \ref{fig:motionplanning}, a three-layer motion planning framework is carefully designed, which consists of global path planning, local path optimization, and velocity planning. The global path planner is employed to provide a rough route from the current robot pose to the goal one, and the local path optimization combined with time-optimal velocity planning is used to generate safe, smooth, and efficient motion commands according to real-time sensor data. Different from the two-layer motion planning framework, local motion planning is decoupled into local path optimization and velocity planning in this work. Compared with the path/velocity coupled planning approaches such as TEB, the adopted path/velocity decoupled framework can achieve better motion efficiency and computational efficiency while ensuring smoothness. The reason is twofold: 1) Decoupled approaches divide the complex motion planning problem into two subproblems. Compared with the original problem, each subproblem has a lower dimension, and the computational efficiency of solving two subproblems separately is higher than directly solving the original complex problem; 2) Too many constraints need to be considered when jointly planning path and velocity, which may lead to mutual compromises. Based on the above careful consideration, the three-layer motion planning framework is designed to consider different constraints and objectives in different layers to avoid mutual conflicts.

\begin{figure}[t]
	\centering
	\includegraphics[scale=0.11]{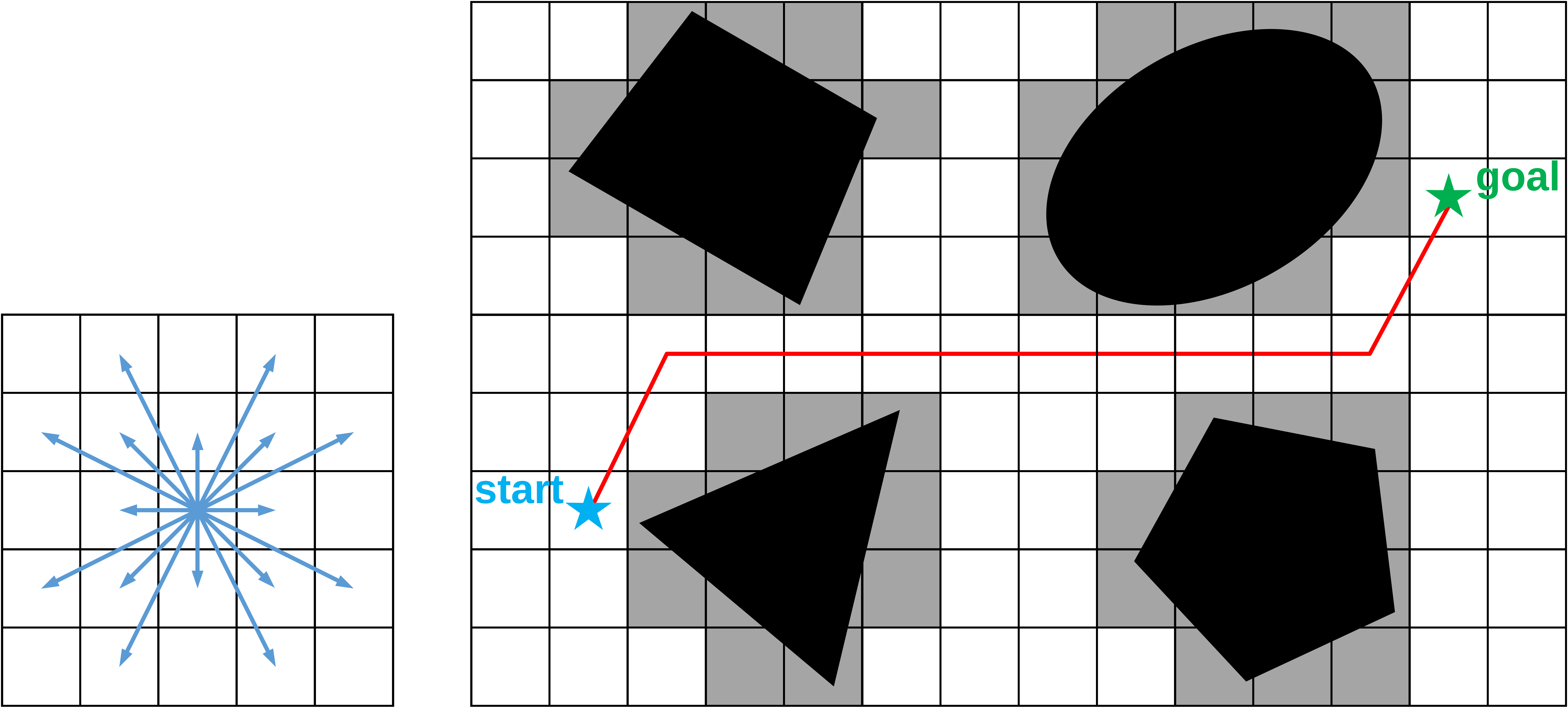}
	\caption{An illustration of the simplified 2-D search in a 16-connected grid. Obstacles are shown in black, and the gray cells denote the dangerous areas around obstacles. An environment-constrained heuristic path is represented by a red polyline, which provides significant guidance for the original 3-D search in complex environments.}
	\label{fig:heuristic2d}
\end{figure}

\section{Heuristic-guided pruning strategy of motion primitives}
To compute a kinematically feasible collision-free global path first, the state lattice-based path planner is employed in this work. A typical state lattice-based path planner consists of graph construction and graph search \cite{likhachev2009planning}. As for graph search, A* is employed in this work. As for graph construction, the edge between two nodes in the graph is built from motion primitives which fulfill the kinematic constraints of the robot. In this section, a novel pruning strategy of motion primitives is proposed to further improve the computational efficiency of the state lattice-based path planner.

\subsection{Problem Formulation}
In this work, we adopt a three-dimensional state representation $ \left( {x,y,\theta } \right) $, where $ \left( {x,y} \right) $ denotes the position of the robot in the world, $ \theta $ represents the heading of the robot and is normalized to $ \left(-\pi, \pi \right] $. For the sake of computational efficiency, the velocities of the robot are not taken into account in the state representation in the global path planning stage. Instead, we temporarily assume that the robot travels at constant linear and angular velocities. Furthermore, in order to obtain a kinematically feasible path, the connectivity between two states is built from motion primitives which fulfill the kinematic constraints of the robot. A motion primitive $ \gamma\left(s, s'\right) $ consists of a sequence of internal robot poses when moving from state $ s $ to state $ s' $. In this work, we follow the popular search-based planning library (SBPL)\footnote{\url{http://wiki.ros.org/sbpl}} to design motion primitives offline, where the trajectory generation algorithm described in \cite{howard2007optimal} is employed with the unicycle model to generate motion primitives for differential-drive robots. If a motion primitive $ \gamma\left(s, s'\right) $ is collision-free, the cost $ g\left( \gamma\left(s, s'\right) \right) $ of this motion primitive is defined as the travel time spent on it. Otherwise, $ g\left( \gamma\left(s, s'\right) \right) $ is set to infinity.

Based on the above preliminaries, the path planning problem is defined as follows. The input of the path planner is an up-to-date EDG, the kinematically feasible motion primitives, the current robot state $ {s_{start}} $ and a goal state $ {s_{goal}} $. And the output is a path that is collision-free and satisfies the kinematic constraints of the robot. Meanwhile, the generated path is expected to be optimal or sub-optimal with respect to the travel time in the state space. 

\subsection{Environment-Constrained 2-D Heuristic}
To cope with the complex environmental constraints, it is usual to solve a simplified search problem online and take the result of the simplified search to guide the original, complex search. In this work, the environment-constrained 2-D heuristic $ h_{2D} $ presented in \cite{likhachev2009planning} is used to guide the A* search. Given the up-to-date environmental information, a 2-D version ($ \left(x, y\right) $) of the original path planning problem is solved by Dijkstra's algorithm. Namely, the original 3-D search problem is reduced to a 2-D search in a 8-connected or 16-connected grid, and the non-holonomic constraints of the robot are ignored. Such a simplified search procedure computes the costs of the shortest paths from the goal cell to other cells in the 2-D grid and stores them as a lookup table. Furthermore, to make the heuristic more informative, the costs of those cells whose distances to obstacles are less than the inscribed radius of the robot footprint are set to infinity. Intuitively, $ h_{2D} $ captures the geometry of obstacles in the environment and guides the more expensive 3-D search away from those areas with dense obstacles. Therefore, $ h_{2D} $ needs to be updated whenever the environment changes. An environment-constrained heuristic path is illustrated in Fig. \ref{fig:heuristic2d}.

\subsection{Motion Primitives Pruning}
\emph{Fully investigating $ h_{2D} $ and designing a pruning strategy of motion primitives to further improve the computational efficiency of graph search is the main novelty of this section.} In \cite{likhachev2009planning}, $ h_{2D} $ is only used to estimate the cost of the shortest path from a given search state to the goal state. In this work, $ h_{2D} $ is further exploited to provide a one-step forward prediction for the search direction, and only part of the motion primitives are involved in each expanding process. Therefore, the branching factor of graph search is decreased, bringing significant benefits for computational efficiency and memory consumption.

\begin{figure*}[t]
	\subfigure[]{\includegraphics[scale=0.23]{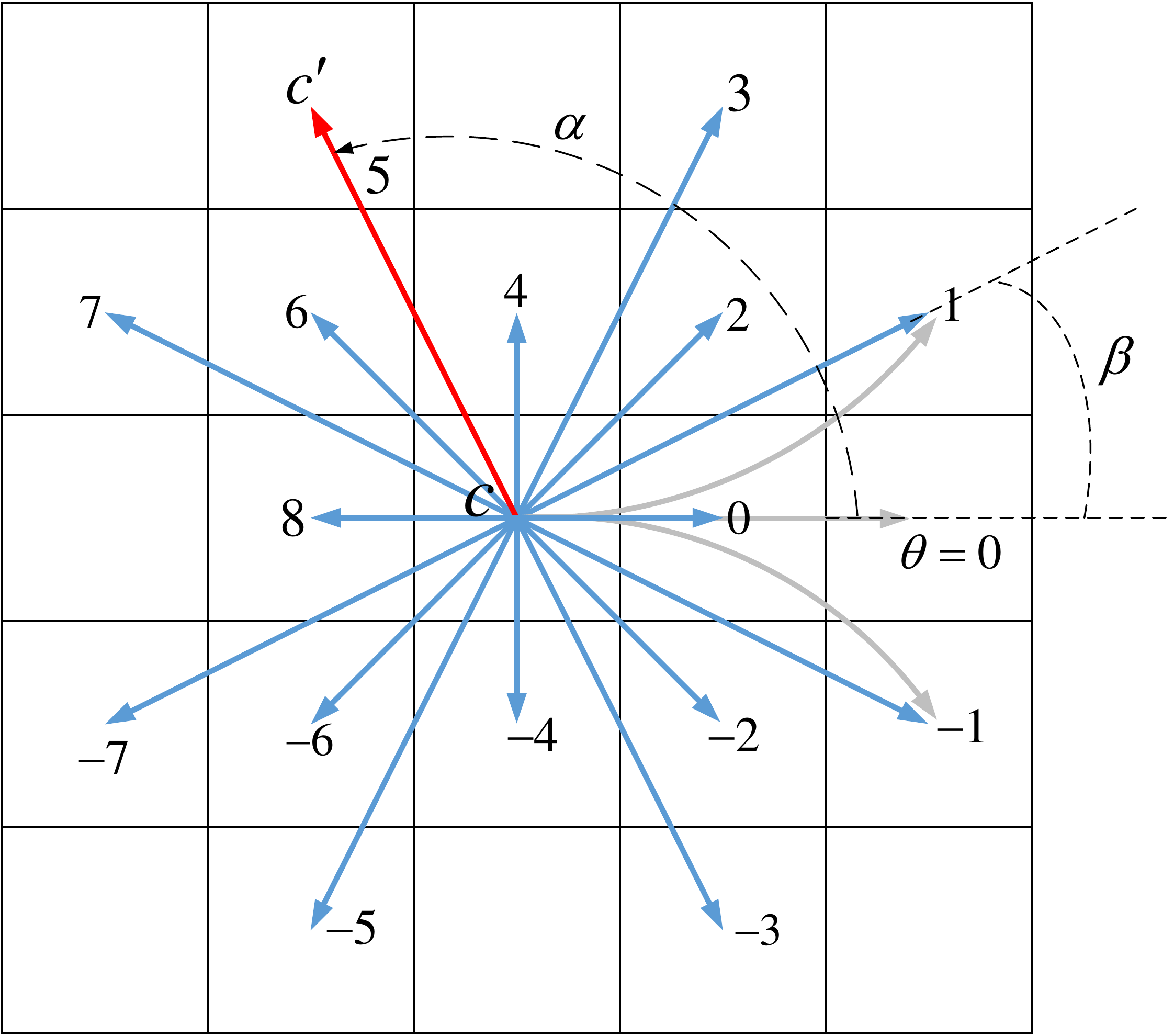}}
	\centering
	\subfigure[]{\includegraphics[scale=0.23]{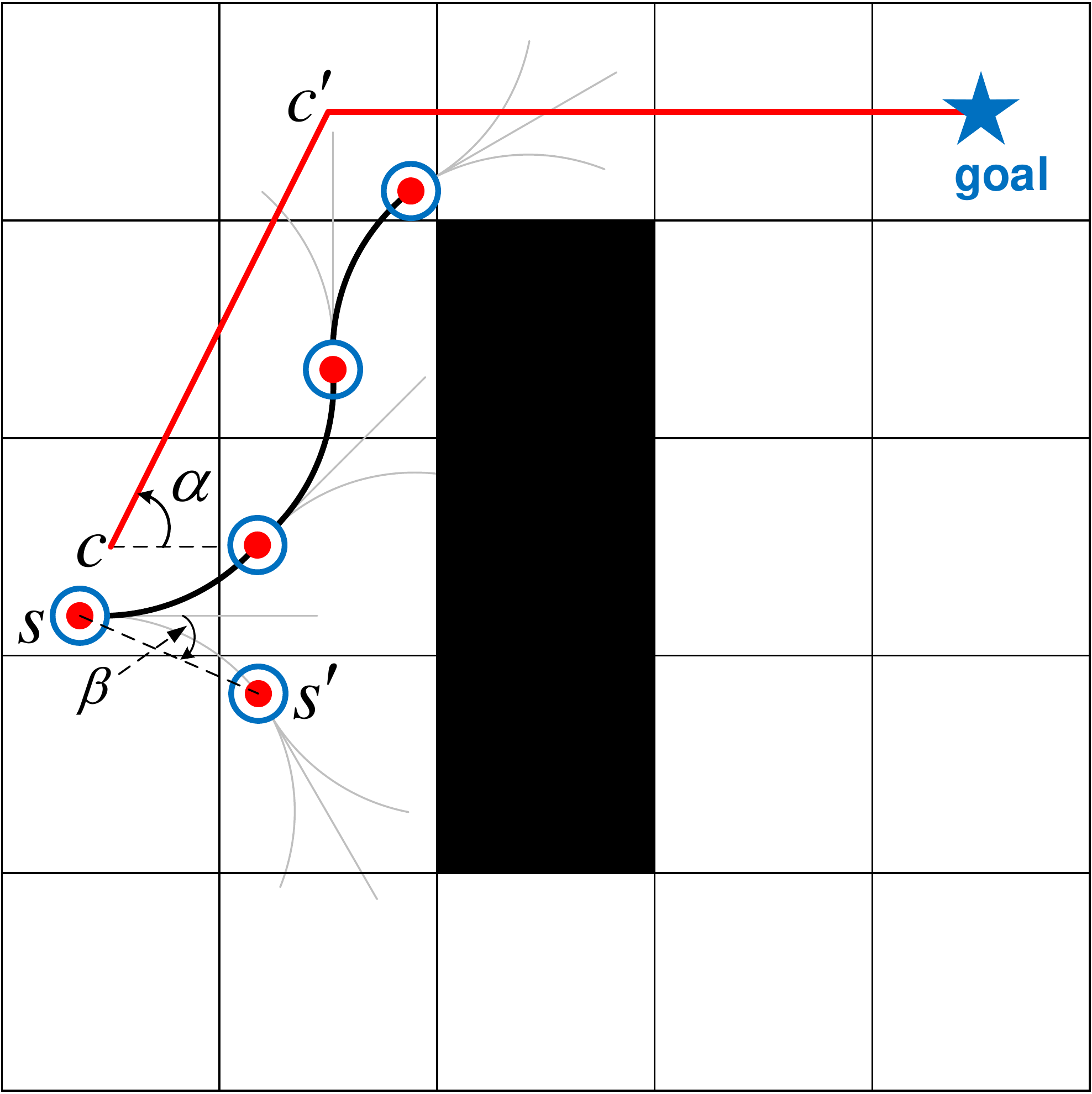}}
	\centering
	\hspace{0.4cm}
	\subfigure[]{\includegraphics[scale=0.23]{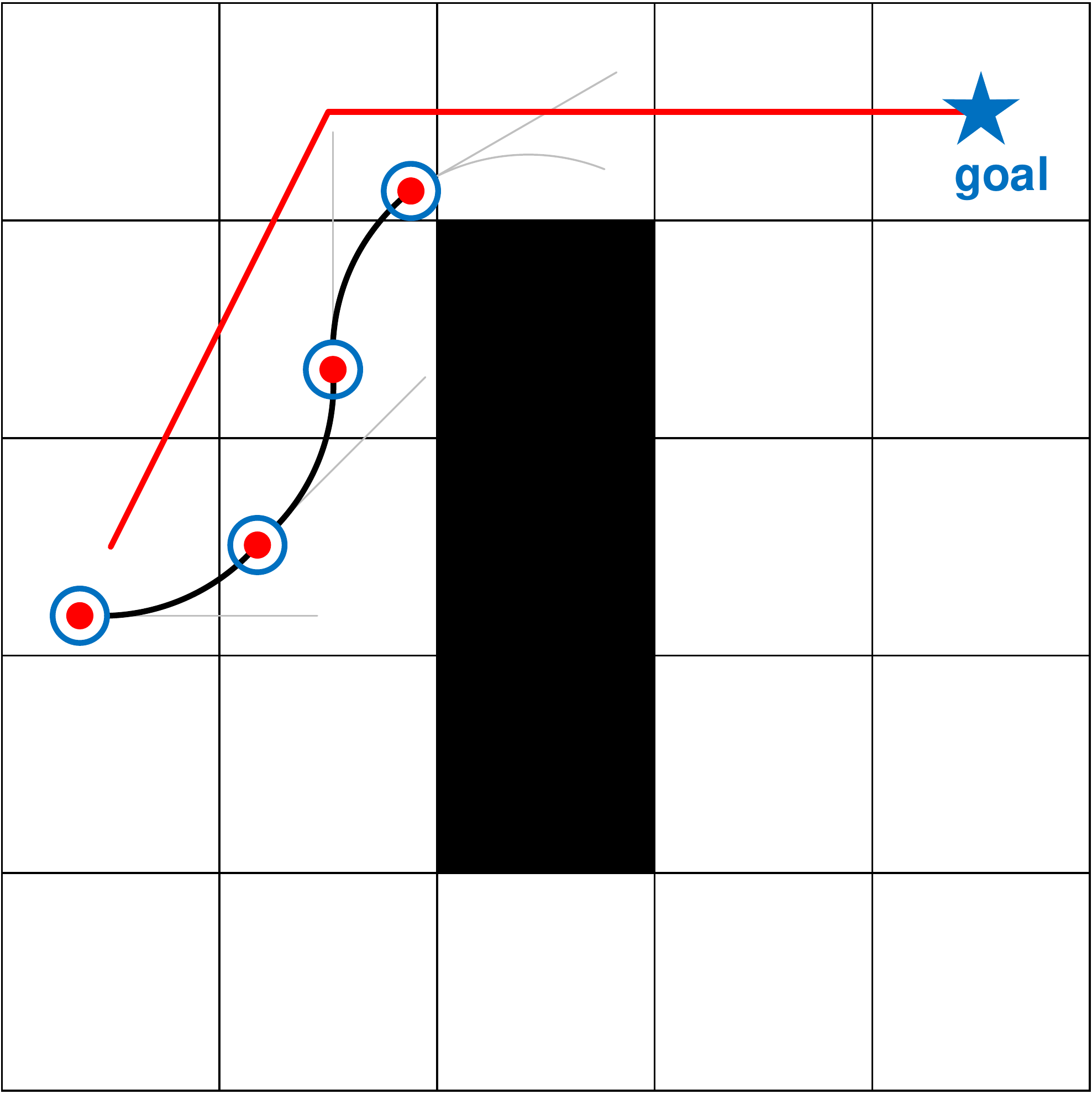}}
	\caption{Illustration of the pruning strategy of motion primitives. Obstacles are shown in black. The red polyline denotes the environment-constrained heuristic path, and the light gray curves represent the designed motion primitives. (a) Orientation for both 3-D search and 2-D search is discretized into 16 angles. (b) Planning with the standard state lattice-based path planner. (c) Planning with the pruning strategy of motion primitives. It should be noted that the start and end points of motion primitives are designed to land on the cell center. To make the illustration clear, we make an appropriate offset.}
	\label{fig:pruning}
\end{figure*}

Above all, it should be noted that the 2-D search is performed in a backward manner, i.e., the start of the 2-D search is the corresponding goal cell of the 3-D search. As illustrated in Fig. \ref{fig:pruning}(a), the orientation space is discretized into 16 angles, and motion primitives from each angle are designed. According to the angular resolution of the 3-D search, the 2-D search is performed correspondingly in a 16-connected grid to construct $ h_{2D} $. The procedure of the proposed pruning strategy of motion primitives is as follows:

\begin{enumerate}[\hspace{1em}1)]
	\item For every cell $ c $ in the 2-D grid except the goal cell, the shortest 2-D path from the goal cell to $ c $ is computed by Dijkstra's algorithm. Furthermore, the predecessor $ c' $ of $ c $ in the 2-D heuristic path is retrieved and the angle $ \alpha $ between the positive $ x $-axis and the vector from $ c $ to $ c' $ is computed and recorded in the cell $ c $.
	\item During the 3-D search stage, for the currently expanded 3-D $ \left(x, y, \theta\right) $ search state $ s $, every one of its successor $ s' $ is retrieved according to the designed motion primitives. The angle $ \beta $ between the positive $ x $-axis and the vector from $ s $ to $ s' $ is computed, as shown in Fig. \ref{fig:pruning}(b). Furthermore, the corresponding 2-D cell $ c $ of $ s $ in the 2-D grid is derived, and the angle $ \alpha $ stored in $ c $ is retrieved. If the deviation between $ \left(\alpha - \theta\right) $ and $ \left(\beta - \theta\right) $ is greater than a threshold $ \varepsilon $, the vector from $ s $ to $ s' $ is considered to point to an unpromising search direction and the corresponding motion primitive is not involved in the actual expanding process. According to the angular resolution of the 3-D search, $ \varepsilon $ is set to $ \frac{\pi }{4} $ in this work.
	\item In the end, only part of the designed motion primitives are involved in each expanding process, as shown in Fig. \ref{fig:pruning}(c).
\end{enumerate}
In the above process, all angles are normalized to $ \left( { - \pi ,\pi } \right] $. The positive sign of the angle corresponds to the counterclockwise direction. It should be noted that three basic motion primitives, namely, ``taking a step forward'', ``turning in place left'', and ``turning in place right'', are always involved in each expanding process regardless of the pruning strategy. The reason is that at least one feasible path can be obtained by only combining these three basic actions if a physically feasible solution indeed exists in the discrete search space. Therefore, the resolution-completeness of the newly proposed path planner will not be lost due to the extension of the pruning strategy to the state lattice-based path planner.

Intuitively, the 2-D cell $ c $ and its predecessor $ c' $ in the environment-constrained heuristic path provide a one-step forward prediction for the 3-D search direction. Based on the pruning strategy, the branching factor of graph search is decreased dramatically and the computational efficiency is significantly improved. Theoretically, we cannot guarantee the optimality of the proposed path planner with motion primitives pruning. However, the practical performance is satisfactory, which will be demonstrated through extensive simulations and experiments in Sections VI and VII.

\section{Soft-constrained local path optimization with sparse-banded system structure}
The path generated by the global path planner is kinematically feasible, but it is still piecewise linear and not suitable for velocity planning. In addition, the global path planner updates with a relatively low frequency. Therefore, the safety of the path needs to be further improved by the real-time sensor data. 

In this section, a new soft-constrained local path optimization approach is proposed to deform an initial path generated by the global path planner. The sparse-banded system structure of the underlying optimization problem is fully investigated and the LM algorithm combined with the forward elimination and back substitution algorithm is utilized to efficiently solve the problem.

\subsection{Problem Formulation}
Given the world coordinates of $ N $ path vertices $ \mathbf{x}_i = {\left( {x_i}, {y_i} \right)}^{\mathrm{T}}, 1 \le i \le N $, a multi-objective path optimization formulation is defined as
\begin{equation}
	\begin{aligned}
		f\left( {\mathbf{x}} \right) 
		&= {\omega _s}\sum\limits_{i = 2}^{N - 1} {{{\left( {{\mathbf{\Delta}} {{\mathbf{x}}_{i + 1}} - {\mathbf{\Delta}} {{\mathbf{x}}_i}} \right)}^{\mathrm{T}}}\left( {{\mathbf{\Delta}} {{\mathbf{x}}_{i + 1}} - {\mathbf{\Delta}} {{\mathbf{x}}_i}} \right)} \\ 
		&+ {\omega _o}\sum\limits_{i = 1}^N {f_o^2\left( {{{\mathbf{x}}_i}} \right)},
	\end{aligned}
	\label{eq1}
\end{equation}
and the optimal parameter vector $ {{\mathbf{x}}^ * } $ are obtained by solving the non-linear least squares problem
\begin{equation}
	{{\mathbf{x}}^ * } = \mathop {\arg \min }\limits_{\mathbf{x}} f\left( {\mathbf{x}} \right).
	\label{eq2}
\end{equation}
Here, $ {\mathbf{x}} = {[{\mathbf{x}}_1^{\mathrm{T}}\;{\mathbf{x}}_2^{\mathrm{T}}\; \ldots \;{\mathbf{x}}_N^{\mathrm{T}}]^{\mathrm{T}}} $ is a $ 2N $-dimensional parameter vector, $ {\mathbf{\Delta}} {{\mathbf{x}}_i} = {{\mathbf{x}}_i} - {{\mathbf{x}}_{i - 1}},2 \le i \le N $ denotes the displacement vector at the vertex $ {{\mathbf{x}}_i} $, and $ {\omega _s} $ and $ {\omega _o} $ are the weights of the cost terms. $ {f_o}\left( {{{\mathbf{x}}_i}} \right) $ is a piecewise continuous differentiable cost function with $ {d_s} $ specifying the minimum safety distance to obstacles:
\begin{equation}
	{f_o}\left( {{{\mathbf{x}}_i}} \right) = 
	\left\{
	\begin{aligned}
		&{d_s} - \tau\left( {{{\mathbf{x}}_i}} \right) &\tau\left( {{{\mathbf{x}}_i}} \right) < {d_s}\\
		&0 &\tau\left( {{{\mathbf{x}}_i}} \right) \ge {d_s}
	\end{aligned}
	\right.,
\end{equation}
wherein $ \tau\left( {{{\mathbf{x}}_i}} \right) $ represents the Euclidean distance between the path vertex $ {{{\mathbf{x}}_i}} $ and the closest obstacle.

\begin{figure}[t]
	\centering
	\includegraphics[scale=0.78]{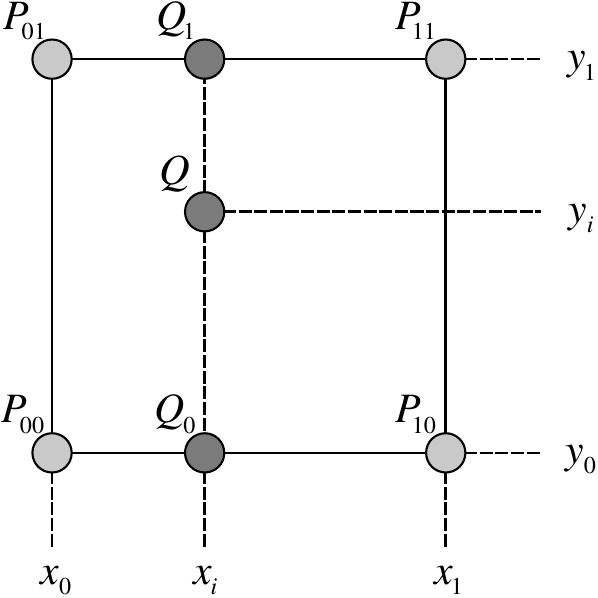}
	\caption{Illustration of bilinear interpolation. Bilinear interpolation first performs linear interpolation along the $ x $-axis and then does linear interpolation along the $ y $-axis. $ Q_0 $ and $ Q_1 $ are the intermediate results of bilinear interpolation in the $ x $-axis.}
	\label{fig:bilinear}
\end{figure}

The first term in Eq. \eqref{eq1} is a measure of the smoothness of the local path. The cost function $ {{\mathbf{\Delta}} {{\mathbf{x}}_{i + 1}} - {\mathbf{\Delta}} {{\mathbf{x}}_i}} $ can be rewritten as $ {{\mathbf{F}}_{i + 1,i}} + {{\mathbf{F}}_{i - 1,i}} $, where $ {{\mathbf{F}}_{i + 1,i}} = {{\mathbf{x}}_{i + 1}} - {{\mathbf{x}}_i} $ and $ {{\mathbf{F}}_{i - 1,i}} = {{\mathbf{x}}_{i - 1}} - {{\mathbf{x}}_i} $. From a physical point of view, this cost term treats the local path as a series spring system, where $ {{\mathbf{F}}_{i + 1,i}} + {{\mathbf{F}}_{i - 1,i}} $ is the resultant force of two springs connecting the vertices $ {{\mathbf{x}}_{i + 1}} $, $ {{\mathbf{x}}_i} $ and $ {{\mathbf{x}}_{i - 1}} $, $ {{\mathbf{x}}_i} $, respectively. If all the resultant forces are equal to zero, all the vertices would uniformly distribute in a straight line, and the local path is ideally smooth.

The second term in Eq. \eqref{eq1} is a measure of the safety of the local path, which is efficiently evaluated in a grid-interpolation scheme. Firstly, an EDG is built by performing distance transform on top of an occupancy grid, wherein each cell of EDG stores the Euclidean distance $ {f_d}\left( P \right) $ between the center $ P $ of this cell and the center of the closest cell occupied by obstacles. Secondly, considering the discrete nature of EDG, which limits the precision of the evaluation of safety and does not allow the direct computation of derivatives, bilinear interpolation is employed to approximate gradients \cite{kohlbrecher2011flexible}. As depicted in Fig. \ref{fig:bilinear}, given the world coordinates $ {{\mathbf{x}}_i} = {({x_i},{y_i})^{\mathrm{T}}} $ of a path vertex $ Q $, $ \tau\left( {{{\mathbf{x}}_i}} \right) $, i.e., the measure of the distance between $ Q $ and the closest obstacle, is approximated as
\begin{equation}
\begin{aligned}
\tau\left( {{{\mathbf{x}}_i}} \right)
\approx &\frac{{{y_1} - {y_i}}}{{{y_1} - {y_0}}}\left( {\frac{{{x_1} - {x_i}}}{{{x_1} - {x_0}}}{f_d}\left( {{P_{00}}} \right) + \frac{{{x_i} - {x_0}}}{{{x_1} - {x_0}}}{f_d}\left( {{P_{10}}} \right)} \right)\\
+ &\frac{{{y_i} - {y_0}}}{{{y_1} - {y_0}}}\left( {\frac{{{x_1} - {x_i}}}{{{x_1} - {x_0}}}{f_d}\left( {{P_{01}}} \right) + \frac{{{x_i} - {x_0}}}{{{x_1} - {x_0}}}{f_d}\left( {{P_{11}}} \right)} \right),
\end{aligned}
\end{equation}
where $ {P_{00}} $, $ {P_{10}} $, $ {P_{01}} $, and $ {P_{11}} $ denote the centers of four closest cells surrounding $ Q $, and $ \left( {{x_0},{y_0}} \right) $, $ \left( {{x_1},{y_0}} \right) $, $ \left( {{x_0},{y_1}} \right) $, and $ \left( {{x_1},{y_1}} \right) $ are the world coordinates of $ {P_{00}} $, $ {P_{10}} $, $ {P_{01}} $, and $ {P_{11}} $ respectively. The gradient $ \nabla \tau\left( {{{\mathbf{x}}_i}} \right) = \left[ {\frac{{\partial \tau}}{{\partial x}}\left( {{{\mathbf{x}}_i}} \right)\;\frac{{\partial \tau}}{{\partial y}}\left( {{{\mathbf{x}}_i}} \right)} \right] $ is approximated as 
\begin{equation}
\begin{aligned}
\frac{{\partial \tau}}{{\partial x}}\left( {{{\mathbf{x}}_i}} \right)
\approx &\frac{{{y_1} - {y_i}}}{{\left( {{x_1} - {x_0}} \right)\left( {{y_1} - {y_0}} \right)}}\left( {{f_d}\left( {{P_{10}}} \right) - {f_d}\left( {{P_{00}}} \right)} \right)\\
+ &\frac{{{y_i} - {y_0}}}{{\left( {{x_1} - {x_0}} \right)\left( {{y_1} - {y_0}} \right)}}\left( {{f_d}\left( {{P_{11}}} \right) - {f_d}\left( {{P_{01}}} \right)} \right),
\end{aligned}
\end{equation}
\begin{equation}
\begin{aligned}
\frac{{\partial \tau}}{{\partial y}}\left( {{{\mathbf{x}}_i}} \right)
\approx &\frac{{{x_1} - {x_i}}}{{\left( {{x_1} - {x_0}} \right)\left( {{y_1} - {y_0}} \right)}}\left( {{f_d}\left( {{P_{01}}} \right) - {f_d}\left( {{P_{00}}} \right)} \right)\\
+ &\frac{{{x_i} - {x_0}}}{{\left( {{x_1} - {x_0}} \right)\left( {{y_1} - {y_0}} \right)}}\left( {{f_d}\left( {{P_{11}}} \right) - {f_d}\left( {{P_{10}}} \right)} \right),
\end{aligned}
\end{equation}
Based on the above grid-interpolation scheme, the safety of the local path is efficiently evaluated while the problem of discretization is overcome.

\emph{Remark:}  Eq. \eqref{eq2} has a solution if the initial guess of Eq. \eqref{eq2} is feasible and the weight of the safety constraint is dominant.

If the weight of the safety constraint is dominant, the LM algorithm will try to minimize the safety term as much as possible, and each path vertex $ \mathbf{x}_i $ will be adjusted independently along the direction of the negative gradient of $ {f_o}\left( \mathbf{x}_i \right) $. Namely, each path vertex will be pushed away from obstacles until the zero gradient occurs.  For $ {f_o}\left( \mathbf{x}_i \right) $, the zero gradient occurs in the safe areas where the distance to the closest obstacle is greater than $ d_s $. Therefore, $ \mathbf{x} $ will converge to a ``safer'' solution than the initial guess. If the initial guess is feasible, the solution of Eq. \eqref{eq2} is guaranteed to be feasible.

\subsection{Least Squares Optimization}
For simplicity of notation, in the rest of this paper we take
\begin{equation}
	\begin{aligned}
		{{\mathbf{S}}_{i - 1,i,i + 1}}\left( {\mathbf{x}} \right) &\coloneqq {\mathbf{\Delta }}{{\mathbf{x}}_{i + 1}} - {\mathbf{\Delta }}{{\mathbf{x}}_i},  \\
		{{\mathbf{O}}_i}\left( {\mathbf{x}} \right) &\coloneqq {f_o}\left( {{{\mathbf{x}}_i}} \right),
	\end{aligned}
\end{equation}
and Eq. \eqref{eq1} is rewritten as 
\begin{equation}
	\begin{aligned}
		f\left( {\mathbf{x}} \right) 
		&= {\omega _s}\sum\limits_{i = 2}^{N - 1} {{{\mathbf{S}}_{i - 1,i,i + 1}}{{\left( {\mathbf{x}} \right)}^{\mathrm{T}}}{{\mathbf{S}}_{i - 1,i,i + 1}}\left( {\mathbf{x}} \right)}  \\
		&+ {\omega _o}\sum\limits_{i = 1}^N {{{\mathbf{O}}_i}{{\left( {\mathbf{x}} \right)}^{\mathrm{T}}}{{\mathbf{O}}_i}\left( {\mathbf{x}} \right)}.
	\end{aligned}
	\label{eq8}
\end{equation}

Assume that a good initial guess $ {\breve{\mathbf{x}}} $ of the parameter vector $ \mathbf{x} $ is known, a numerical solution of Eq. \eqref{eq2} can be obtained by using the
popular Gauss-Newton or LM algorithms. The idea is to approximate the cost term by its first order Taylor expansion around the current initial guess
\begin{equation}
	\begin{aligned}
		{{\mathbf{S}}_{i - 1,i,i + 1}}\left( {{\breve{\mathbf{x}}} + {\mathbf{\Delta x}}} \right) &\approx {{\mathbf{S}}_{i - 1,i,i + 1}} + {\mathbf{J}}_{i - 1,i,i + 1}^s{\mathbf{\Delta x}}, \\	
		{{\mathbf{O}}_i}\left( {\breve{\mathbf{x}} + {\mathbf{\Delta x}}} \right) &\approx {{\mathbf{O}}_i} + {\mathbf{J}}_i^o{\mathbf{\Delta x}},
	\end{aligned}
	\label{eq9}
\end{equation}
where $ {\mathbf{J}}_{i - 1,i,i + 1}^s $ and $ {\mathbf{J}}_i^o $ are the Jacobians of $ {{\mathbf{S}}_{i - 1,i,i + 1}}\left( {\mathbf{x}} \right) $ and $ {{\mathbf{O}}_i}\left( {\mathbf{x}} \right) $ computed in $ \breve{\mathbf{x}} $, respectively. For simplicity of notation we take $ {{\mathbf{S}}_{i - 1,i,i + 1}} \coloneqq {{\mathbf{S}}_{i - 1,i,i + 1}}\left( {\breve{\mathbf{x}}} \right) $
and $ {{\mathbf{O}}_i} \coloneqq {{\mathbf{O}}_i}\left( \breve{\mathbf{x}} \right) $. Substituting Eq. \eqref{eq9} in the cost terms in Eq. \eqref{eq8}, we obtain
\begingroup\makeatletter\def\f@size{9}\check@mathfonts
\def\maketag@@@#1{\hbox{\m@th\normalsize\normalfont#1}}%
\begin{equation}
	\begin{aligned}
		&\ {{{\mathbf{S}}_{i - 1,i,i + 1}}{{\left( {{\breve{\mathbf{x}}} + {\mathbf{\Delta x}}} \right)}^{\mathrm{T}}}{{\mathbf{S}}_{i - 1,i,i + 1}}\left( {{\breve{\mathbf{x}}} + {\mathbf{\Delta x}}} \right)}\\
		&\approx {\left( {{{\mathbf{S}}_{i - 1,i,i + 1}} + {\mathbf{J}}_{i - 1,i,i + 1}^s{\mathbf{\Delta x}}} \right)^{\mathrm{T}}}\left( {{{\mathbf{S}}_{i - 1,i,i + 1}} + {\mathbf{J}}_{i - 1,i,i + 1}^s{\mathbf{\Delta x}}} \right) \\
		&= \underbrace {{\mathbf{S}}_{i - 1,i,i + 1}^{\mathrm{T}}{{\mathbf{S}}_{i - 1,i,i + 1}}}_{c_{i - 1,i,i + 1}^s} + 2\underbrace {{\mathbf{S}}_{i - 1,i,i + 1}^{\mathrm{T}}{\mathbf{J}}_{i - 1,i,i + 1}^s}_{{{\left( {{\mathbf{b}}_{i - 1,i,i + 1}^s} \right)}^{\mathrm{T}}}}{\mathbf{\Delta x}} \\
		&+ {\mathbf{\Delta }}{{\mathbf{x}}^{\mathrm{T}}}\underbrace {{{\left( {{\mathbf{J}}_{i - 1,i,i + 1}^s} \right)}^{\mathrm{T}}}\left( {{\mathbf{J}}_{i - 1,i,i + 1}^s} \right)}_{{\mathbf{H}}_{i - 1,i,i + 1}^s}{\mathbf{\Delta x}}  \\
		&= c_{i - 1,i,i + 1}^s + 2{{\left( {{\mathbf{b}}_{i - 1,i,i + 1}^s} \right)}^{\mathrm{T}}}{\mathbf{\Delta x}} + {\mathbf{\Delta }}{{\mathbf{x}}^{\mathrm{T}}}{\mathbf{H}}_{i - 1,i,i + 1}^s{\mathbf{\Delta x}},
	\end{aligned}
\end{equation}
\endgroup
and
\begingroup\makeatletter\def\f@size{9}\check@mathfonts
\def\maketag@@@#1{\hbox{\m@th\normalsize\normalfont#1}}%
\begin{equation}
	\begin{aligned}
		&\ {{\mathbf{O}}_i}{\left( {{\breve{\mathbf{x}}} + {\mathbf{\Delta x}}} \right)^{\mathrm{T}}}{{\mathbf{O}}_i}\left( {{\breve{\mathbf{x}}} + {\mathbf{\Delta x}}} \right)\\
		&\approx {\left( {{{\mathbf{O}}_i} + {\mathbf{J}}_i^o{\mathbf{\Delta x}}} \right)^{\mathrm{T}}}\left( {{{\mathbf{O}}_i} + {\mathbf{J}}_i^o{\mathbf{\Delta x}}} \right)\\
		&= \underbrace {{\mathbf{O}}_i^{\mathrm{T}}{{\mathbf{\Omega }}_o}{{\mathbf{O}}_i}}_{c_i^o} + 2\underbrace {{\mathbf{O}}_i^{\mathrm{T}}{{\mathbf{\Omega }}_o}{\mathbf{J}}_i^o}_{{{\left( {{\mathbf{b}}_i^o} \right)}^{\mathrm{T}}}}{\mathbf{\Delta x}} + {\mathbf{\Delta }}{{\mathbf{x}}^{\mathrm{T}}}\underbrace {{{\left( {{\mathbf{J}}_i^o} \right)}^{\mathrm{T}}}{{\mathbf{\Omega }}_o}\left( {{\mathbf{J}}_i^o} \right)}_{{\mathbf{H}}_i^o}{\mathbf{\Delta x}}\\
		&= c_i^o + 2{\left( {{\mathbf{b}}_i^o} \right)^{\mathrm{T}}}{\mathbf{\Delta x}} + {\mathbf{\Delta }}{{\mathbf{x}}^{\mathrm{T}}}{\mathbf{H}}_i^o{\mathbf{\Delta x}}.
	\end{aligned}
\end{equation}
\endgroup

With the above local approximations, we can linearize the function given in Eq. \eqref{eq8} around the current initial guess
\begingroup\makeatletter\def\f@size{9}\check@mathfonts
\def\maketag@@@#1{\hbox{\m@th\normalsize\normalfont#1}}%
\begin{equation}
	\begin{aligned}
		&\ f\left( {{\breve{\mathbf{x}}} + {\mathbf{\Delta x}}} \right)\\
		&= {\omega _s}\sum\limits_{i = 2}^{N - 1} {{{\mathbf{S}}_{i - 1,i,i + 1}}{{\left( {{\breve{\mathbf{x}}} + {\mathbf{\Delta x}}} \right)}^{\mathrm{T}}}{{\mathbf{S}}_{i - 1,i,i + 1}}\left( {{\breve{\mathbf{x}}} + {\mathbf{\Delta x}}} \right)} \\
		&+ {\omega _o}\sum\limits_{i = 1}^N {{{\mathbf{O}}_i}{{\left( {{\breve{\mathbf{x}}} + {\mathbf{\Delta x}}} \right)}^{\mathrm{T}}}{{\mathbf{O}}_i}\left( {{\breve{\mathbf{x}}} + {\mathbf{\Delta x}}} \right)} \\
		&\approx {\omega _s}\sum\limits_{i = 2}^{N - 1} {c_{i - 1,i,i + 1}^s + 2{{\left( {{\mathbf{b}}_{i - 1,i,i + 1}^s} \right)}^{\mathrm{T}}}{\mathbf{\Delta x}} + {\mathbf{\Delta }}{{\mathbf{x}}^{\mathrm{T}}}{\mathbf{H}}_{i - 1,i,i + 1}^s{\mathbf{\Delta x}}} \\
		&+ {\omega _o}\sum\limits_{i = 1}^N {c_i^o + 2{{\left( {{\mathbf{b}}_i^o} \right)}^{\mathrm{T}}}{\mathbf{\Delta x}} + {\mathbf{\Delta }}{{\mathbf{x}}^{\mathrm{T}}}{\mathbf{H}}_i^o{\mathbf{\Delta x}}} \\
		&= {c^s} + 2{\left( {{{\mathbf{b}}^s}} \right)^{\mathrm{T}}}{\mathbf{\Delta x}} + {\mathbf{\Delta }}{{\mathbf{x}}^{\mathrm{T}}}{{\mathbf{H}}^s}{\mathbf{\Delta x}}\\
		&+ {c^o} + 2{\left( {{{\mathbf{b}}^o}} \right)^{\mathrm{T}}}{\mathbf{\Delta x}} + {\mathbf{\Delta }}{{\mathbf{x}}^{\mathrm{T}}}{{\mathbf{H}}^o}{\mathbf{\Delta x}}\\
		&= c + 2{{\mathbf{b}}^{\mathrm{T}}}{\mathbf{\Delta x}} + {\mathbf{\Delta }}{{\mathbf{x}}^{\mathrm{T}}}{\mathbf{H\Delta x}},
	\end{aligned}
	\label{eq12}
\end{equation}
\endgroup
where
\begin{equation}
	\begin{array}{l}
		{c^s} = {\omega _s}\sum {c_{i - 1,i,i + 1}^s},\quad  {c^o} = {\omega _o}\sum {c_i^o,} \\
		{{\mathbf{b}}^s} = {\omega _s}\sum {{\mathbf{b}}_{i - 1,i,i + 1}^s} ,\quad {{\mathbf{b}}^o} = {\omega _o}\sum {{\mathbf{b}}_i^o} ,\\
		{{\mathbf{H}}^s} = {\omega _s}\sum {{\mathbf{H}}_{i - 1,i,i + 1}^s} , \quad {{\mathbf{H}}^o} = {\omega _o}\sum {{\mathbf{H}}_i^o} ,\\
		c = {c^s} + {c^o},\quad {\mathbf{b}} = {{\mathbf{b}}^s} + {{\mathbf{b}}^o},\quad {\mathbf{H}} = {{\mathbf{H}}^s} + {{\mathbf{H}}^o}.
	\end{array}
\end{equation}

Eq. \eqref{eq12} can be minimized in $ {{\mathbf{\Delta x}}} $ by taking the derivative of $ f\left( {{\breve{\mathbf{x}}} + {\mathbf{\Delta x}}} \right) $ with respective to $ {{\mathbf{\Delta x}}} $ and setting the result to zero
\begin{equation}
	{\mathbf{H\Delta }}{{\mathbf{x}}^ * } =  - {\mathbf{b}},
	\label{eq14}
\end{equation}
where $ {\mathbf{H}} $ is the system matrix of the optimization problem. The solution of the path optimization problem is obtained by adding the increment $ {\mathbf{\Delta }}{{\mathbf{x}}^ * } $ to the initial guess
\begin{equation}
	{{\mathbf{x}}^ * } = {\breve{\mathbf{x}}} + {\mathbf{\Delta }}{{\mathbf{x}}^ * }.
	\label{eq15}
\end{equation}

The Gauss-Newton algorithm iterates the linearization in Eq. \eqref{eq12}, the solution in Eq. \eqref{eq14}, and the update step in Eq. \eqref{eq15}. In every iteration, the previous solution is used as the linearization point and the initial guess until a given termination criterion is met. The LM algorithm introduces a damping factor and backup actions to Gauss-Newton to control the convergence. Instead of solving Eq. \eqref{eq14}, the LM algorithm solves a damped version
\begin{equation}
	\left(\mathbf{H} + \lambda \mathbf{I}\right) \mathbf{\Delta} \mathbf{x^*} = -\mathbf{b},
	\label{eq16}
\end{equation}
where $ \lambda $ is a damping factor to control the step size in case of nonlinear surfaces.

\begin{figure*}[t]
	\centering
	\includegraphics[scale=0.046]{fig/sparsebandedmatrix.pdf}
	\caption{Superposition construction process of the sparse-banded Hessian matrix of the path optimization problem. Top and bottom matrices denote the Hessian matrices of the smoothness and safety constraints, respectively.}
	\label{fig:bandedmatrix}
\end{figure*}

\subsection{Sparse-Banded System Structure}
An important property of the underlying optimization problem is the sparse-banded structure of the system matrix $ {\mathbf{H}} $ 
\begingroup\makeatletter\def\f@size{9}\check@mathfonts
\def\maketag@@@#1{\hbox{\m@th\normalsize\normalfont#1}}%
\begin{equation}
	\begin{aligned}
		{\mathbf{H}} &= {{\mathbf{H}}^s} + {{\mathbf{H}}^o}\\
		&= {\omega _s}\sum\limits_{i = 2}^{N - 1} {{\mathbf{H}}_{i - 1,i,i + 1}^s\left( {\breve{\mathbf{x}}} \right)}  + {\omega _o}\sum\limits_{i = 1}^N {{\mathbf{H}}_i^o\left( {\breve{\mathbf{x}}} \right)}  \\
		&= {\omega _s}\sum\limits_{i = 2}^{N - 1} {{{\left( {{\mathbf{J}}_{i - 1,i,i + 1}^s} \right)}^{\mathrm{T}}}\left( {{\mathbf{J}}_{i - 1,i,i + 1}^s} \right)}  + {\omega _o}\sum\limits_{i = 1}^N {{{\left( {{\mathbf{J}}_i^o} \right)}^{\mathrm{T}}}\left( {{\mathbf{J}}_i^o} \right)}.
	\end{aligned}
\end{equation}
\endgroup
Recall that $ {{\mathbf{S}}_{i - 1,i,i + 1}}\left( {\mathbf{x}} \right) = {\mathbf{\Delta }}{{\mathbf{x}}_{i + 1}} - {\mathbf{\Delta }}{{\mathbf{x}}_i} = {{\mathbf{x}}_{i + 1}} - 2{{\mathbf{x}}_i} + {{\mathbf{x}}_{i - 1}} $ is a function of the three consecutive variables $ {{{\mathbf{x}}_{i - 1}}} $, $ {{{\mathbf{x}}_i}} $, and $ {{{\mathbf{x}}_{i + 1}}} $. Therefore, the partial derivatives of the variables other than these three variables in $ {\mathbf{J}}_{i - 1,i,i + 1}^s\left( {\breve{\mathbf{x}}} \right) $ are all zero
\begingroup\makeatletter\def\f@size{9}\check@mathfonts
\def\maketag@@@#1{\hbox{\m@th\normalsize\normalfont#1}}%
\begin{equation}
	\setlength{\arraycolsep}{1.2pt}
	{\mathbf{J}}_{i - 1,i,i + 1}^s\left( {\breve{\mathbf{x}}} \right) = \left( {\begin{array}{*{20}{c}}
			{\mathbf{0}}, &\cdots, &{\mathbf{0}}, &{{{\mathbf{I}}_{2 \times 2}}}, &{ - 2{{\mathbf{I}}_{2 \times 2}}}, &{{{\mathbf{I}}_{2 \times 2}}}, &{\mathbf{0}}, & \cdots, &{\mathbf{0}}
	\end{array}} \right),
\end{equation}
\endgroup
and $ {\mathbf{H}}_{i - 1,i,i + 1}^s\left( {\breve{\mathbf{x}}} \right) = {\mathbf{J}}_{i - 1,i,i + 1}^s{\left( {\breve{\mathbf{x}}} \right)^{\mathrm{T}}}{\mathbf{J}}_{i - 1,i,i + 1}^s\left( {\breve{\mathbf{x}}} \right) $ only contributes a $ 6 \times 6 $ diagonal block to $ {\mathbf{H}} $
\begingroup\makeatletter\def\f@size{9}\check@mathfonts
\def\maketag@@@#1{\hbox{\m@th\normalsize\normalfont#1}}%
\begin{equation}
	\setlength{\arraycolsep}{1.2pt}
	{\mathbf{H}}_{i - 1,i,i + 1}^s\left( {\breve{\mathbf{x}}} \right) = \left( {\begin{array}{ccccc}
			\ddots &{}&{}&{}&{}\\
			{}&{{{\mathbf{I}}_{2 \times 2}}}&{ - 2{{\mathbf{I}}_{2 \times 2}}}&{{{\mathbf{I}}_{2 \times 2}}}&{}\\
			{}&{ - 2{{\mathbf{I}}_{2 \times 2}}}&{4{{\mathbf{I}}_{2 \times 2}}}&{ - 2{{\mathbf{I}}_{2 \times 2}}}&{}\\
			{}&{{{\mathbf{I}}_{2 \times 2}}}&{ - 2{{\mathbf{I}}_{2 \times 2}}}&{{{\mathbf{I}}_{2 \times 2}}}&{}\\
			{}&{}&{}&{}& \ddots 
	\end{array}} \right).
\end{equation}
\endgroup
For simplicity of notation we omit the zero blocks. There is a similar property for $ {\mathbf{J}}_i^o\left( {\breve{\mathbf{x}}} \right) $ and $ {{\mathbf{H}}_i^o\left( {\breve{\mathbf{x}}} \right)} $. In the end, $ {\mathbf{H}} $ is a $ 2N \times 2N $ banded matrix with bandwidth 5, as illustrated in Fig. \ref{fig:bandedmatrix}.

\emph{Remark:} Considering a $ n \times n $ banded matrix $ {\mathbf{A}} = ({a_{ij}}) $, the bandwidth of $ {\mathbf{A}} $ is the maximum number $ k $ such that $ {a_{i,j}} = 0 $ if $ \left| {i - j} \right| > k $.
 
For simplicity of notation,  Eq. \eqref{eq16} is rewritten as
\begin{equation}
	\mathbf{A \Delta x^*} = -{\mathbf{b}},
	\label{eq20}
\end{equation}
where $ {\mathbf{A}} = {\mathbf{H}} + \lambda {\mathbf{I}} $. According to the LM algorithm, $ \mathbf{A} $ is symmetric positive-definite. To solve Eq. \eqref{eq20} efficiently, $ \mathbf{A} $ is decomposed as $ {\mathbf{A}} = {\mathbf{LU}} $ by LU decomposition
\begin{equation}
	\mathbf{A\Delta x^*} = {\mathbf{LU\Delta x^*}} = -{\mathbf{b}},
\end{equation}
where $ {\mathbf{L}} $ is a \emph{unit lower triangular} matrix and $ {\mathbf{U}} $ is an \emph{upper triangular} matrix. We can obtain $ \mathbf{\Delta x^*} $ by first solving
\begin{equation}
	\mathbf{Ly} = -{\mathbf{b}},
	\label{eq22}
\end{equation}
and then solving
\begin{equation}
	\mathbf{U \Delta x^*} = {\mathbf{y}}.
	\label{eq23}
\end{equation}
Eq. \eqref{eq22} can be solved by forward elimination since $ {\mathbf{L}} $ is unit lower triangular. To solve Eq. \eqref{eq23}, we can use back substitution since $ {\mathbf{U}} $ is upper triangular. The forward elimination and back substitution algorithm is essentially the Gaussian elimination algorithm, but it makes full use of the non-zero element distribution characteristics of the sparse-banded matrix and significantly reduces the computational complexity. The time complexity of the forward elimination and back substitution algorithm is $ O({k^2} \cdot n) $, while that of the Gaussian elimination algorithm is $ O({n^3}) $, where $ n $ and $ k $ are the size and bandwidth of the banded matrix respectively. In addition, the forward elimination and back substitution algorithm is substantially faster than a general sparse factorization because it avoids having to store the factored form of the matrix. Readers can refer to \cite{datta2010numerical} for more details about the forward elimination and back substitution algorithm.

\subsection{Velocity Profile Generation}
After path optimization, a smooth local path is obtained, but it is still piecewise linear and not suitable for velocity planning. To address this problem, the local path is further smoothed via cubic spline interpolation. Finally, a numerical integration (NI)-based time-optimal velocity planning algorithm presented in \cite{zhang2018multilevel} is employed to generate a feasible linear velocity profile along the smoothed local path. The NI-based algorithm can acquire provably time-optimal trajectory with low computational complexity \cite{pham2014general, shen2017essential, shen2018complete, shen2020real}, which solves the problem by computing maximum velocity curve (MVC) considering both kinematic and environmental constraints and then performing numerical integration under MVC. Readers can refer to \cite{zhang2018multilevel} for more details about the proofs of feasibility, completeness, and time-optimality of this algorithm.

\section{Implementation details}
\subsection{Setup}
E\ensuremath{^3}MoP is implemented in C/C++. The maximum number of iterations of the LM algorithm is set to $ 100 $. The initial guess of local path optimization is obtained by sampling in the global path with an interval of $ 0.1 $ $ \mathrm{m} $. In consideration of the computational efficiency and sensing range, the cumulative length of the initial local path is set to $ 3 $ $ \mathrm{m} $. EDG is computed through an efficient distance transform algorithm described in \cite{felzenszwalb2012distance}. We maintain two EDGs of different resolutions for the hierarchical motion planning framework, wherein $ 0.1 $ $ \mathrm{m/cell} $ for the global grid map and $ 0.05 $ $ \mathrm{m/cell} $ for the local grid map. The dimension of the global grid map corresponds to the global prior map, while the local grid map is a sliding window whose map center is corresponding to the robot pose and the map dimension is set to $ 8 \times 8 $ $ \mathrm{m^2} $. The safety distance $ {d_s} $ is set to $ 0.5 $ $ \mathrm{m} $. Moreover, the modules of the global path planning and EDG updating run in two independent threads, with cycles of $ 250 $ $ \mathrm{ms} $ and $ 40 $ $ \mathrm{ms} $ respectively. All the simulations and experiments are tested on a laptop with an Intel Core i7-9750H processor and 16 GB RAM. 

To obtain a set of feasible and effective weights for the path optimization formulation, we fix $ \omega_s $ to $ 1 $ firstly and then set an interval for $ \omega_o $. After that, $ \omega_o $ is tuned by dichotomy and visual inspection with the help of the ROS visualization tool RViz. In this work, $ \omega_o $ is set to $ 10 $. In the future, we plan to use machine learning techniques to tune these weights adaptively.

\begin{figure*}[t]
	\centering
	\includegraphics[scale=0.207]{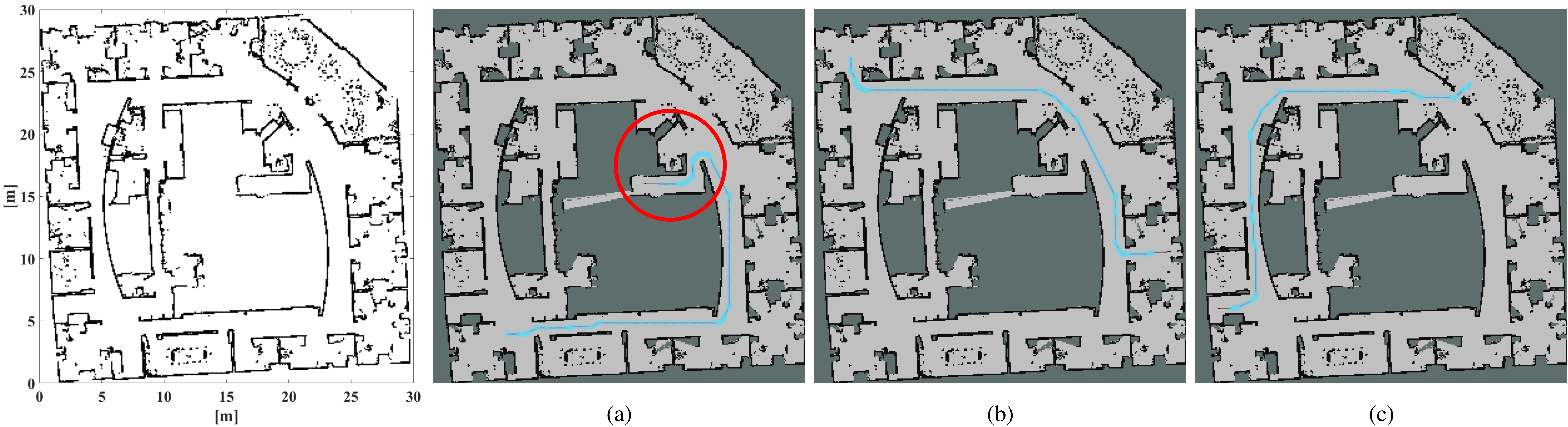}
	\caption{Intel Research Lab. A map derived from real sensor data is used to simulate a 2-D environment. We randomly select three sets of start poses and goal poses to test global path planners, as shown in (a)-(c). The light blue strips indicate the footprints of the robot along the planned paths.}
	\label{fig:intel}
\end{figure*}

\subsection{Metrics}
The proposed pruning strategy of motion primitives is integrated into the standard state lattice-based path planner (A* + motion primitives) \cite{likhachev2009planning} to derive a new path planner (A* + motion primitives + pruning strategy). And the new path planner is compared with the standard state lattice-based path planner to validate the effectiveness of the pruning strategy. Both path planners require offline designed motion primitives. As mentioned before, the trajectory generation algorithm described in \cite{howard2007optimal} is employed with the unicycle model to generate motion primitives for differential-drive robots. Furthermore, the metrics of the number of expanded states and the planning time are used to evaluate the computational efficiency of graph search \cite{likhachev2009planning}, and the graph size, i.e., the number of nodes in the search graph, is employed to evaluate the memory consumption.

To validate the efficiency, smoothness, and flexibility of the proposed optimization-based local planner, we compare it with the advanced quintic B\'{e}zier curve-based state space sampling local planner (QBC) \cite{zhang2018multilevel}. The total travel time is utilized to evaluate the motion efficiency of local planners. In addition, we carefully design various simulation and experimental scenarios to compare the smoothness and flexibility of local planners, which will be detailed in Sections VI and VII.  

\section{Simulations}
In this section, we verify the applicability of the proposed E\ensuremath{^3}MoP in simulation. We choose Stage \cite{vaughan2008massively} as the simulation platform since its lightweight advantage.

\subsection{Simulation Setup}
To make the simulations more realistic and exhibit the real-world cluster, noise, and occlusion effects, the simulation environment is built on top of a map generated from real sensor data. As depicted in Fig. \ref{fig:intel}, we simulate a 2-D environment based on a map built from the Intel Research Lab data set, which is available from the Radish data set repository \cite{howard2003robotics}. The map is constructed by an open-source 2-D laser SLAM system Karto\footnote{\url{http://wiki.ros.org/slam_karto}}. The size of the environment is approximately $ 30 \times 30$ $ \mathrm{m^2} $.

\begin{table}[t]
	\centering
	\caption{Quantitative statistics of path planning results in Intel Research Lab}
	\label{tab:globalplanner}
	\scalebox{0.94}{
		\begin{tabular}{cccccccc}
			\toprule
			\multicolumn{2}{c}{} 	   								& \# of        		& Time     			& Branching  		& Graph     		& Path    	\\
			\multicolumn{2}{c}{} 	     							& expands     		& (secs)      		& factor  			& size   			& cost     	\\
			\midrule
			\multirow{2}{*}{Fig. \ref{fig:intel}(a)} 	& Lattice  	& 56,952        	&  0.052        	& 6.054       		& 77,056        	& 98,830   	\\
														& Ours 		& \textbf{17,657}  	&  \textbf{0.016}  	& \textbf{4.899}   	& \textbf{24,405}   & 98,830    \\
			\midrule
			\multirow{2}{*}{Fig. \ref{fig:intel}(b)}   	& Lattice  	& 54,548        	&  0.049        	& 6.216      		& 78,232        	& 91,738    \\
														& Ours 		& \textbf{17,634}  	&  \textbf{0.013}  	& \textbf{5.132}   	& \textbf{23,917}   & 91,738    \\
			\midrule
			\multirow{2}{*}{Fig. \ref{fig:intel}(c)}   	& Lattice 	& 81,387        	&  0.075        	& 6.057      		& 100,609      		& 108,210  	\\
														& Ours   	& \textbf{30,966}  	&  \textbf{0.024}  	& \textbf{4.788}   	& \textbf{39,600}   & 108,210   \\
			\bottomrule
	\end{tabular}}
\end{table}

\subsection{Comparison on Global Planning}
We randomly select three sets of start poses and goal poses in the simulation environment to test global path planners, as shown in Fig. \ref{fig:intel}(a)-(c). The state lattice-based path planner and the proposed path planner generate equal quality paths. Therefore, we only show one path in each sub-figure. Table \ref{tab:globalplanner} enumerates some quantitative statistics of path planning results in these three sets of simulations.

\subsubsection{Comparison on Computational Efficiency}
Thanks to the pruning strategy of motion primitives, the search direction of the proposed path planner is focused towards the most promising search areas, and the computational efficiency of graph search is significantly improved. Compared with the state lattice-based path planner, the number of expanded states of the proposed path planner decreases by an average of $ 66.21\% $. As a result, planning with the proposed path planner is more than three times faster than planning with the state lattice-based path planner.

\subsubsection{Comparison on Memory Consumption}
Both the state lattice-based path planner and the proposed path planner employ the implicit graph representation. Namely, the memory is allocated according to the need of creating new nodes during the search process, rather than allocating memory for the whole search space in advance. Taking advantage of the pruning strategy of motion primitives, plenty of unpromising search branches are pruned, and the number of the created nodes is reduced. The graph size of the proposed path planner is about $ 33.87\% $ of that of the state lattice-based path planner.

In conclusion, the proposed path planner generates equal quality paths with much less time and memory consumption than the state lattice-based path planner, which demonstrates the effectiveness of the proposed pruning strategy of motion primitives and implicitly validates that the advantage of the pruning strategy is not incorporated in a sophisticated heuristic such as $ h_{2D} $.

\begin{figure}[t]
	\centering
	\includegraphics[scale=0.35]{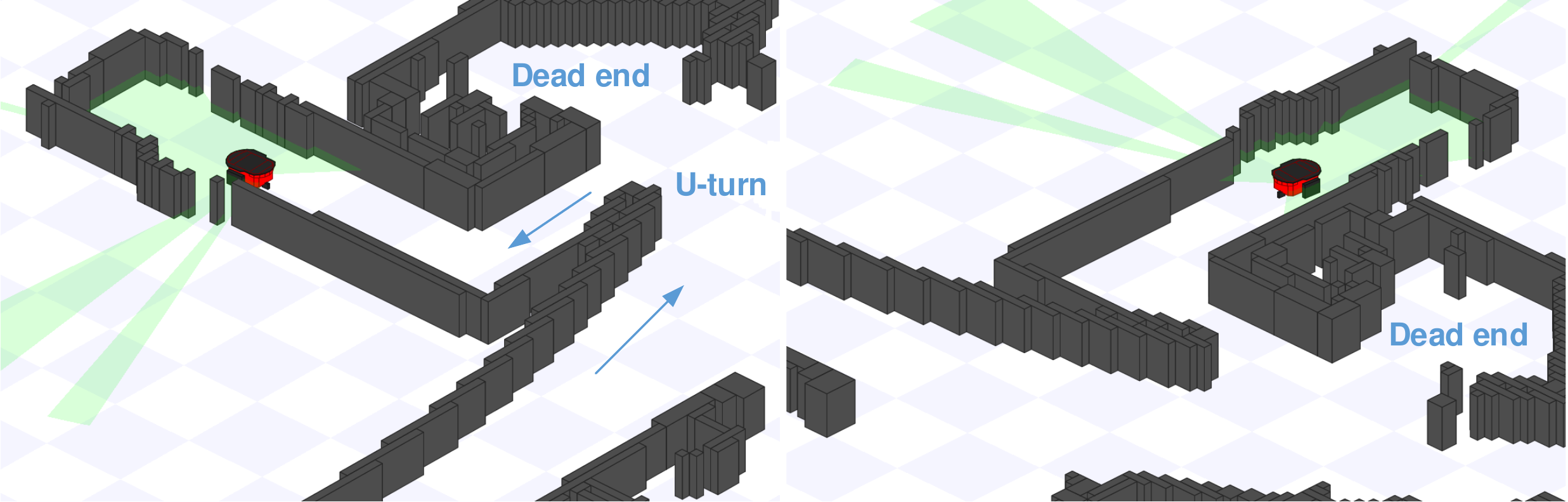}
	\caption{Local details of the simulation environment.}
	\label{fig:uturn}
\end{figure}

\begin{figure}[t]
	\centering
	\subfigure[]{\includegraphics[scale=0.19]{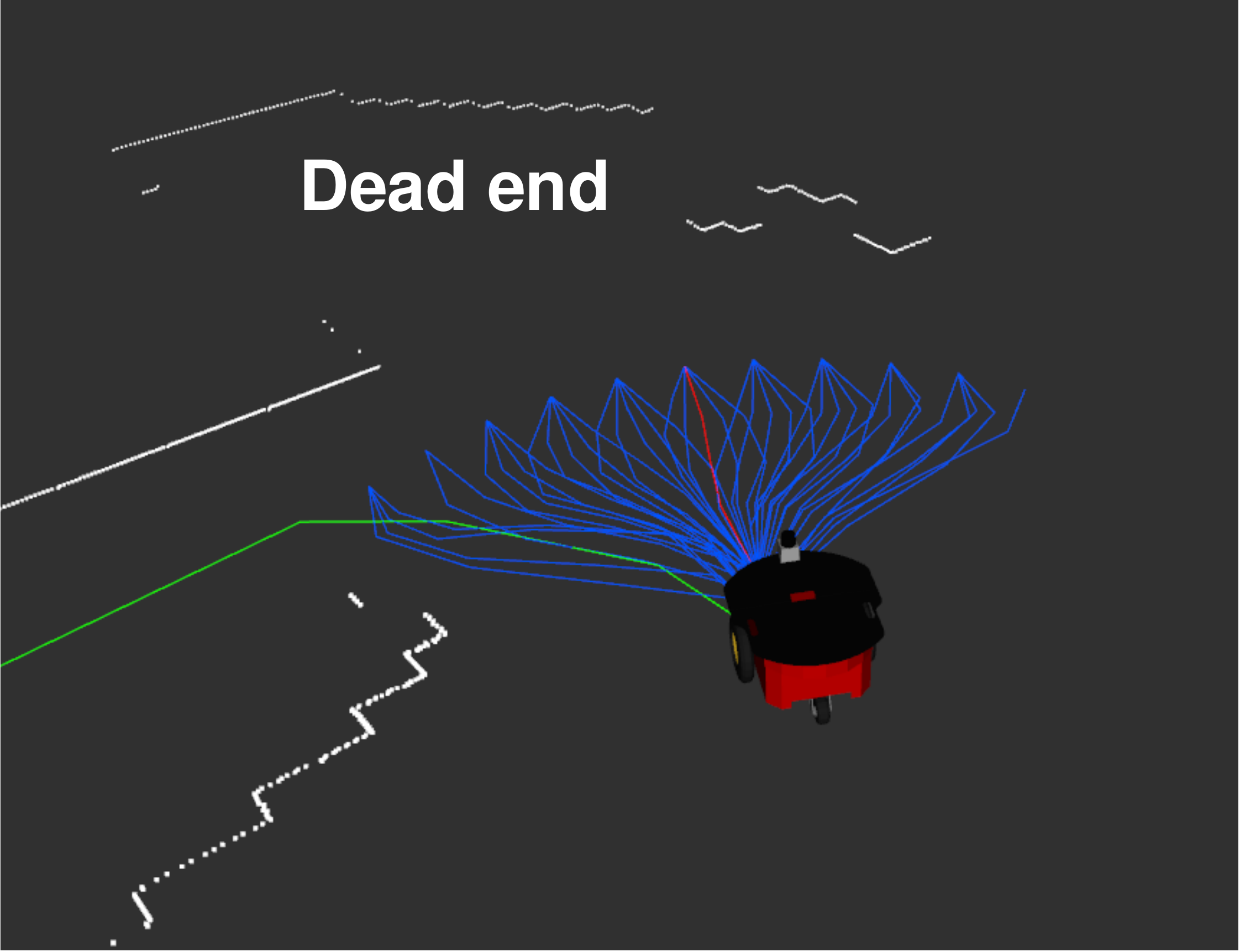}}
	\centering
	\subfigure[]{\includegraphics[scale=0.19]{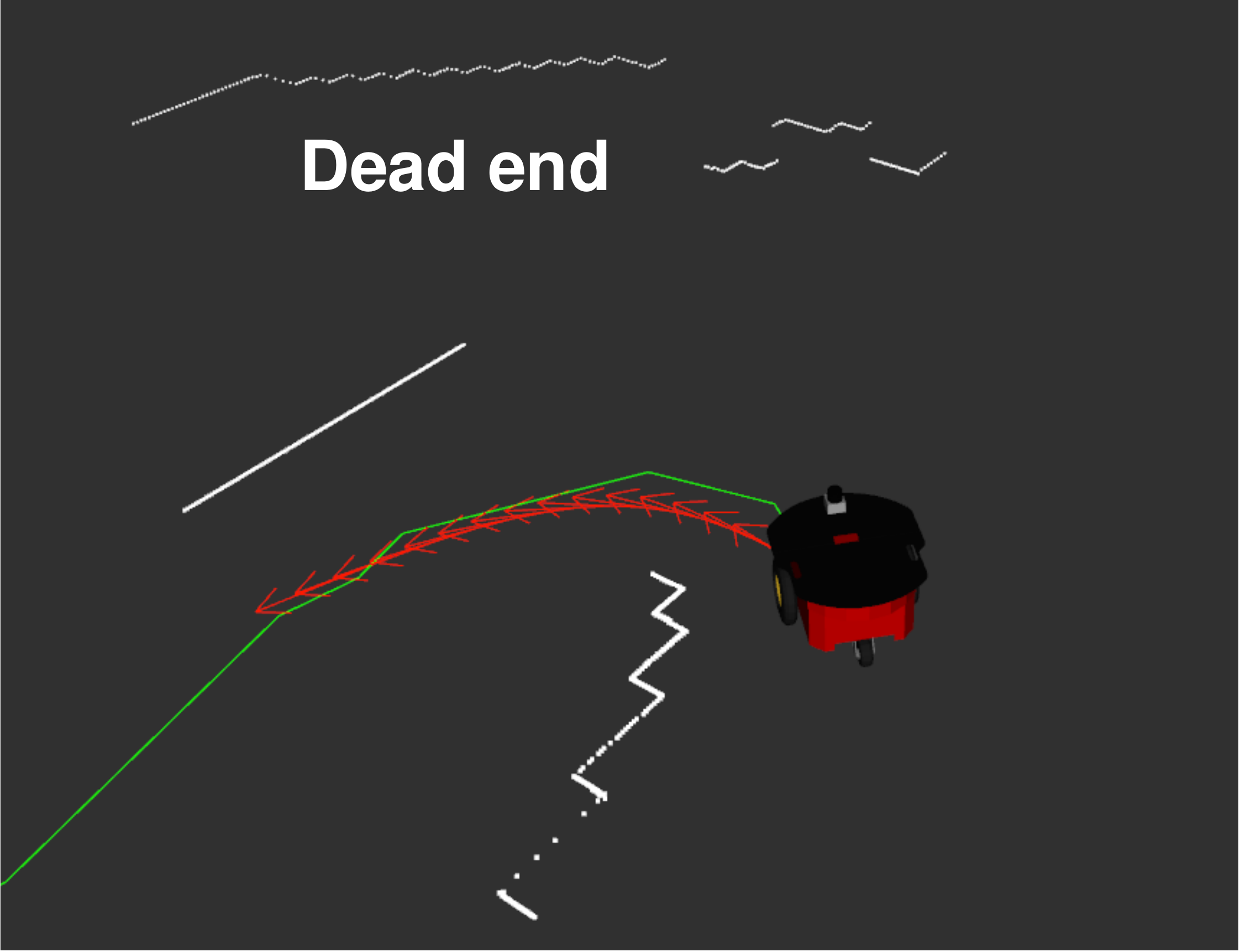}}
	\caption{Comparisons on local planners in simulation. (a) Planning result of QBC. The blue trajectories represent the feasible candidate B\'{e}zier curves and the red trajectory denotes the selected optimal B\'{e}zier curve. (b) Planning result of the proposed local planner. The red arrows indicate the optimized poses. Green trajectories in both (a) and (b) indicate the global paths provided by the global path planner.}
	\label{fig:localplanner}
\end{figure}

\subsection{Comparison on Local Planning}
\subsubsection{Comparison on Motion Flexibility}
To compare the motion flexibility of local planners, we make the robot move along the planned path shown in Fig. \ref{fig:intel}(a). This is an extremely challenging navigation task. Firstly, the robot needs to make a sharp U-turn at the corner, as illustrated in Fig. \ref{fig:intel}(a) and Fig. \ref{fig:uturn}. This process requires local planners to provide flexible motions. Secondly, after going around the corner, the robot needs to go through a narrow polyline corridor to reach the goal, which requires safe motions. \emph{The whole navigation process poses a huge challenge to the flexibility and safety of local planners.}

In the test, QBC guides the robot around the corner slowly. At the turning point of the corner, it selects a going-forward motion instead of a turning-left motion, as depicted in Fig. \ref{fig:localplanner}(a). In the end, the robot comes to a dead end. All candidate paths are infeasible and the local planner fails. On the contrary, the proposed local planner guides the robot around the U-turn smoothly, as shown in Fig. \ref{fig:localplanner}(b). This demonstrates the proposed local planner performs better motion flexibility.

\subsubsection{Comparison on Motion Efficiency}
To compare the motion efficiency of local planners, we make the robot move along the planned paths shown in Fig. \ref{fig:intel}(b) and (c). The local path generated by QBC is made up of many connected pieces of incomplete B\'{e}zier curves and is somewhat rough. On the contrary, the smoothness constraint is explicitly considered in the proposed local path optimization approach. Therefore, the local path generated by the proposed local planner is much smoother than that of QBC. \emph{The performance in smoothness is intuitively reflected in motion efficiency.} In the tests, the proposed local planner takes $ 49.78 $ $ \mathrm{s} $ and $ 50.59 $ $ \mathrm{s} $ respectively to guide the robot to the goals, while QBC costs $ 54.43 $ $ \mathrm{s} $ and $ 55.35 $ $ \mathrm{s} $ respectively under the same conditions. This supports the proposed local planner has higher motion efficiency.

In conclusion, the proposed optimization-based local planner performs better motion flexibility and efficiency than QBC.

\begin{figure}[t]
	\centering
	\subfigure[]{\includegraphics[scale=0.15]{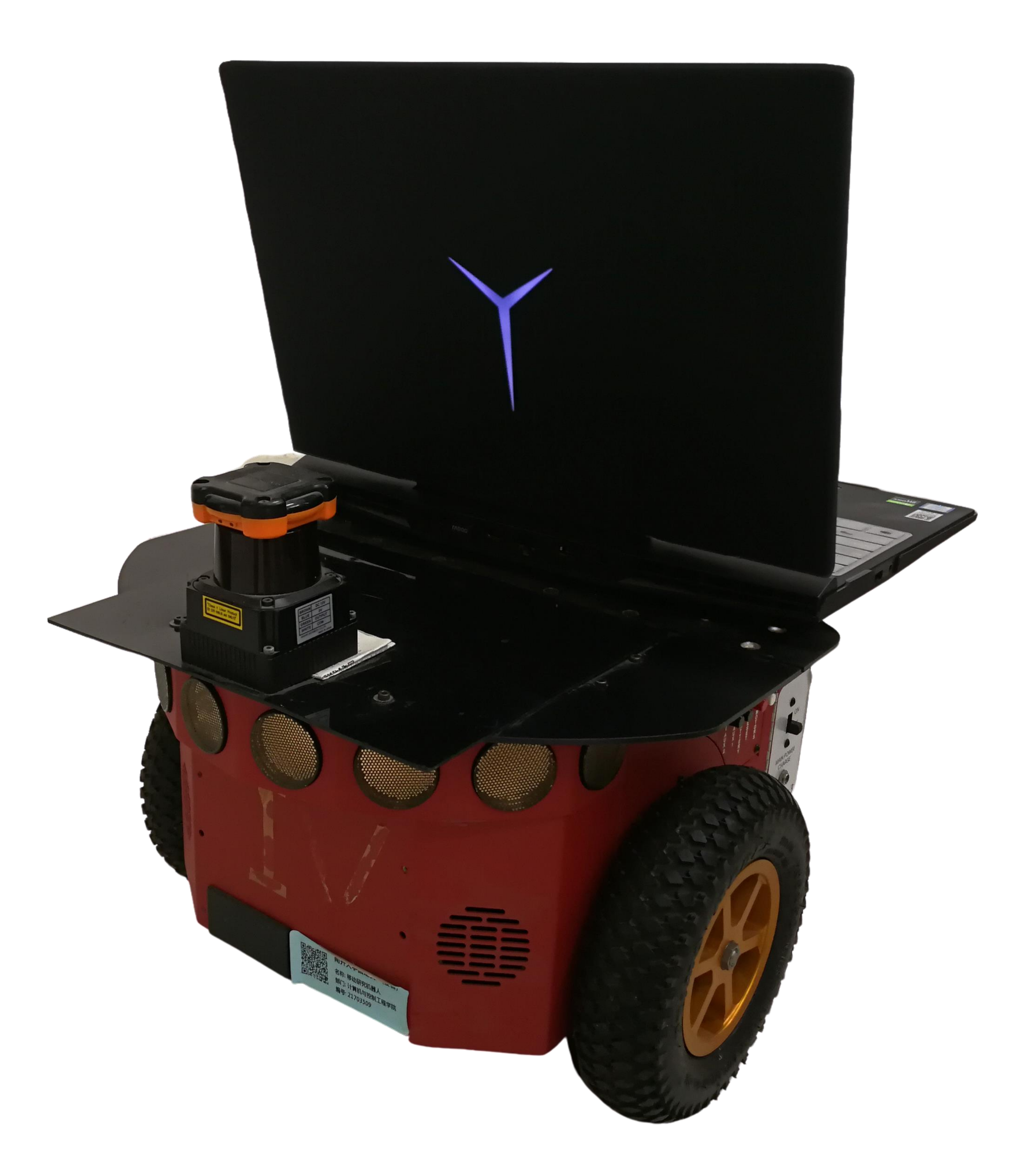}}
	\centering
	\subfigure[]{\includegraphics[scale=0.15]{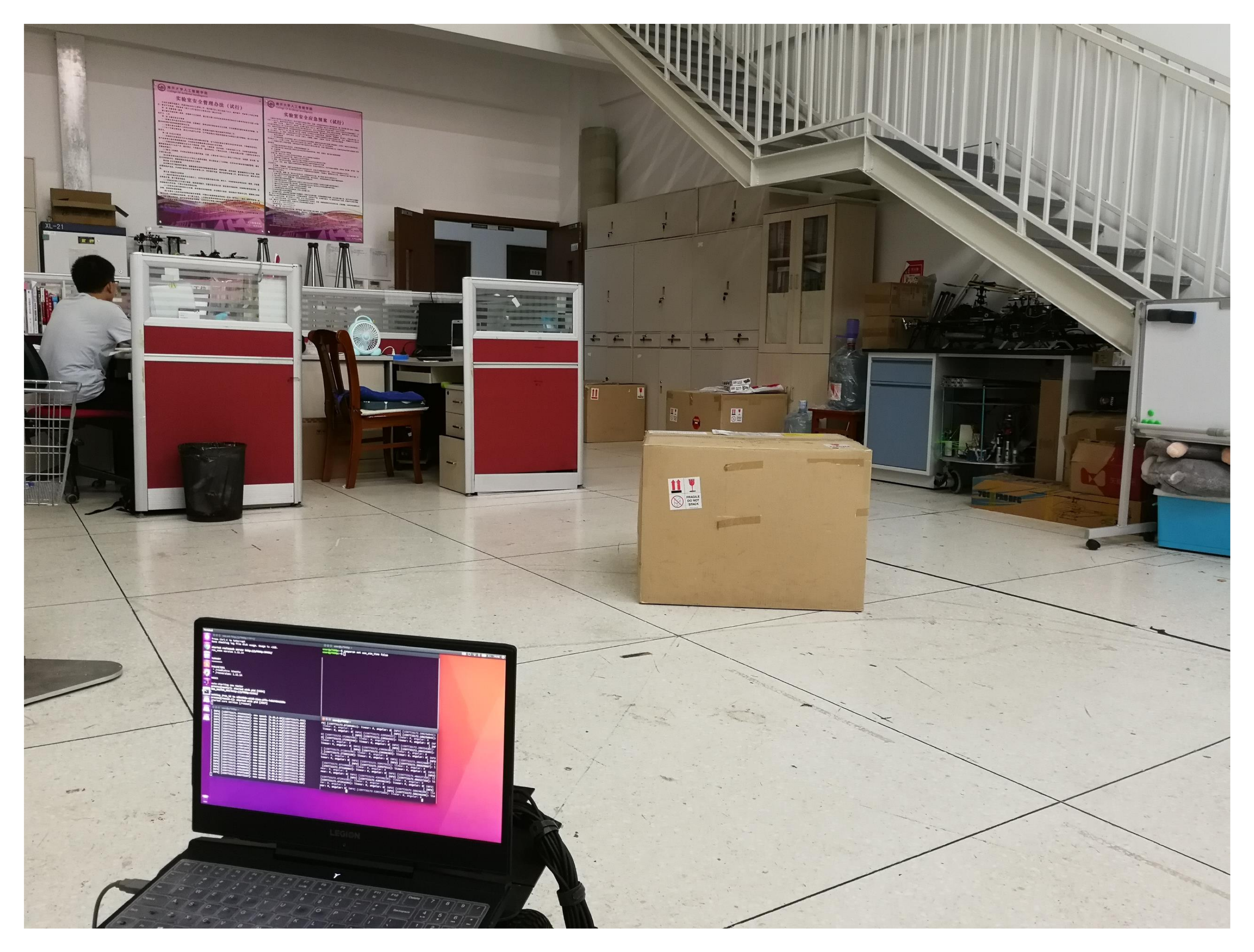}}
	\caption{(a) Experimental platform. (b) Experimental scenario I.}
	\label{fig:platform}
\end{figure}

\section{Experiments}
In this section, we elaborately design three sets of challenging experimental scenarios to verify the efficiency, flexibility, smoothness, and safety of the proposed E\ensuremath{^3}MoP. The experimental results are presented and discussed in detail to validate the superior performance of E\ensuremath{^3}MoP.

\subsection{Experimental Setup}
In this work, the experiments are conducted on a Pioneer 3-DX differential-drive robot equipped with a Hokuyo UTM-30LX laser rangefinder, as shown in Fig. \ref{fig:platform}(a). The maximum linear velocity of the robot is $ 1.2 $ $ \mathrm {m/s} $. Considering the safety of indoor navigation, we set the upper bound of the linear velocity to $ 0.7 $ $ \mathrm {m/s} $. The laser rangefinder has a scanning range of $ {270^ \circ }$ with the angular resolution being $0.25^\circ$, and the effective measurement range is $ 0.1 $ $ \mathrm{m} $ to $ 30 $ $ \mathrm {m} $. Due to the limited angular range, backward motions are not considered in this work. 

\subsection{Comparison on Global Planning}
To validate the superior performance of the proposed path planner, Scenario I is designed as depicted in Fig. \ref{fig:platform}(b). A box is placed in front of the robot as an obstacle. The robot is required to reach the door.

\begin{figure}[t]
	\centering
	\subfigure[]{\includegraphics[scale=0.2]{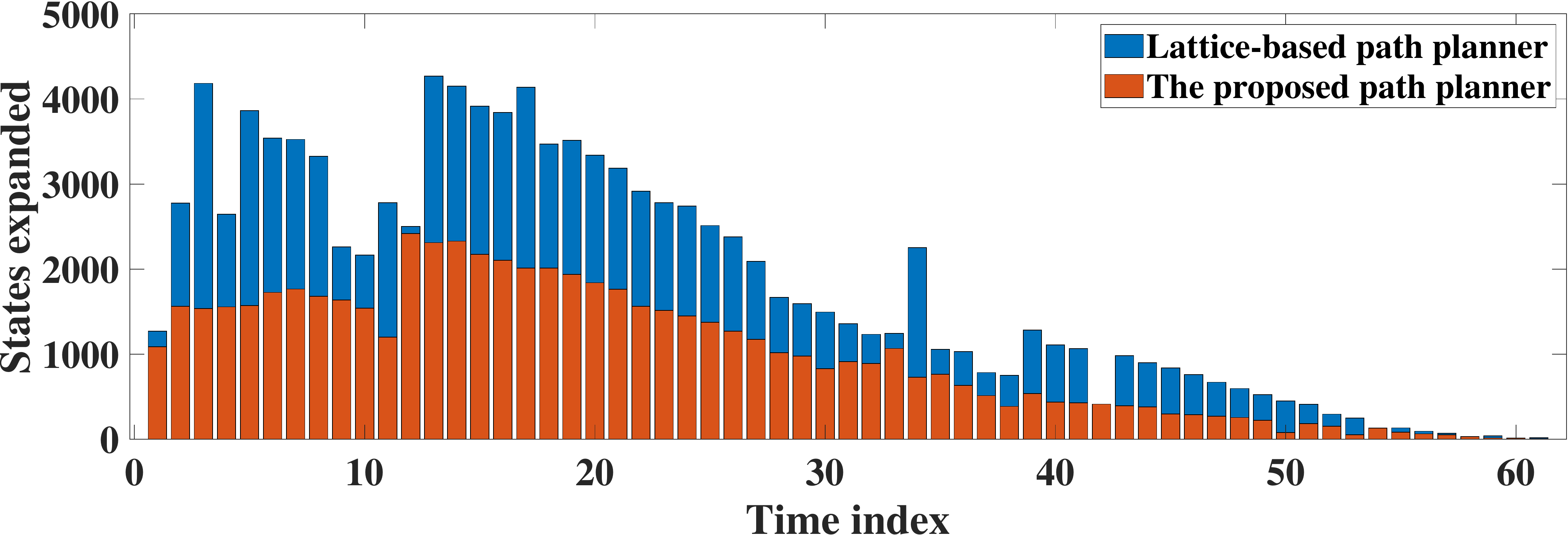}}
	\centering
	\subfigure[]{\includegraphics[scale=0.2]{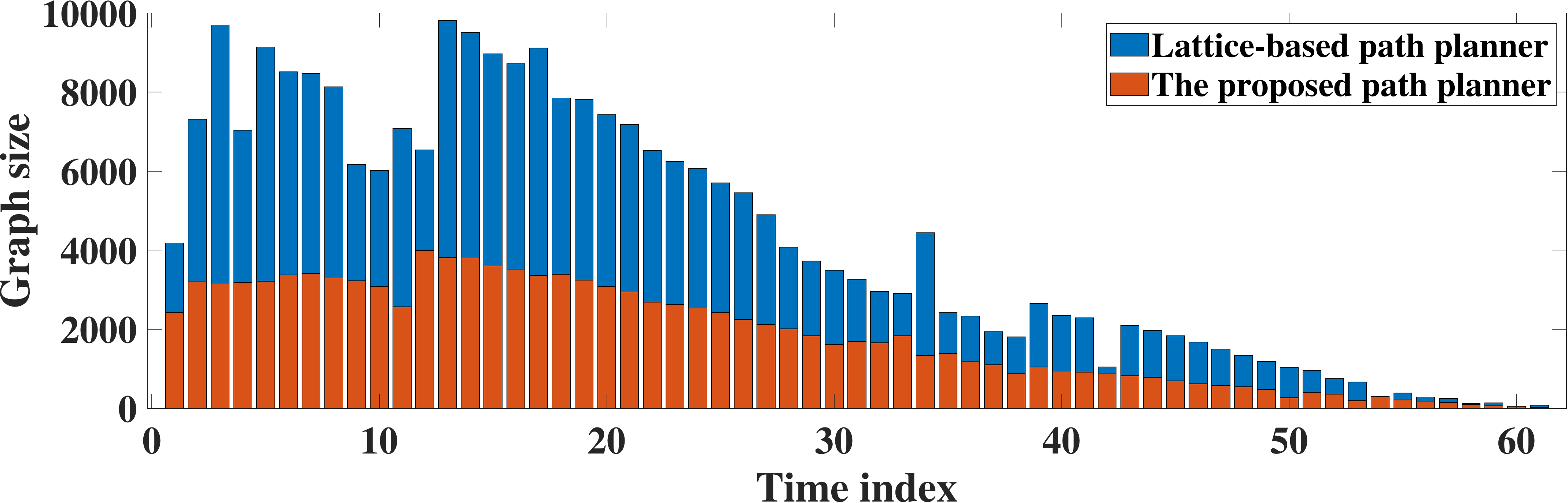}}
	\caption{Comparative experimental results of global path planners in the test of Scenario I. (a) The number of expanded states. (b) The number of nodes in the search graph.}
	\label{fig:navpath}
\end{figure}

To make a fair comparison on global path planners, we use the same local planner in the test of Scenario I. In both sets of tests, the thread of global path planning is triggered $ 61 $ times, and the experimental results are illustrated in Fig. \ref{fig:navpath}. The average number of expanded states and the graph size of the state lattice-based path planner are $ 1794 $ and $ 4226 $ respectively, while those of the proposed path planner are $ 978 $ and $ 1815 $ respectively. Compared with the state lattice-based path planner, the computational efficiency and memory consumption of the proposed path planner are improved by $ 45.48\% $ and $ 57.05\% $ respectively, which demonstrates the effectiveness of the proposed pruning strategy of motion primitives.

\begin{table}[t]
	\centering
	\caption{Comparisons of Motion Efficiency (Seconds) in Scenario II}
	\label{tab:localplanner}
	\begin{tabular}{ccccc}
		\toprule
		& QBC planner \cite{zhang2018multilevel}	& TEB planner \cite{rosmann2012trajectory}	& Ours		\\
		\midrule
		1	& 25.614  & 23.625 	& \textbf{19.813}  			\\
		2	& 25.711  & 23.417	& \textbf{19.842}  			\\
		3	& 25.872  & 24.224 	& \textbf{19.921}  			\\
		4	& 25.836  & 23.626 	& \textbf{19.909} 			\\
		5	& 25.743  & 23.871 	& \textbf{19.834}   		\\
		\bottomrule
	\end{tabular}
\end{table}

\begin{figure}[t]
	\centering
	\subfigure[]{\includegraphics[scale=0.1826]{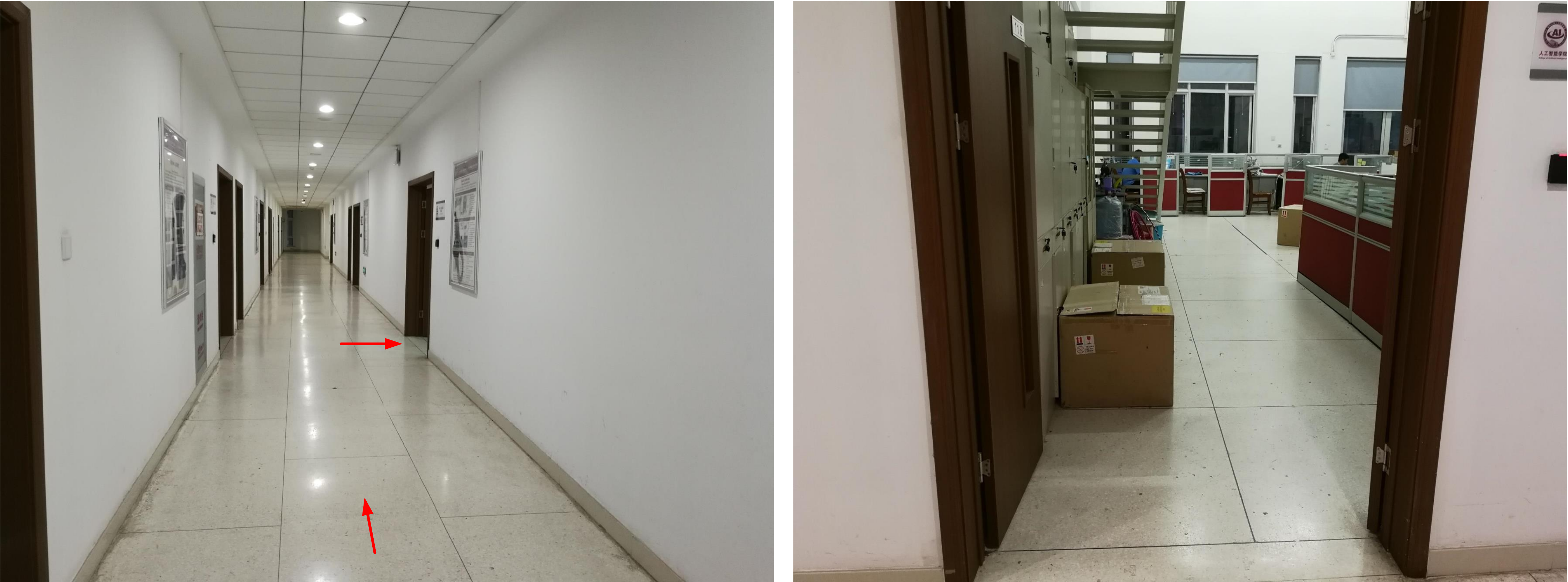}}
	\centering
	\subfigure[]{\includegraphics[scale=0.18]{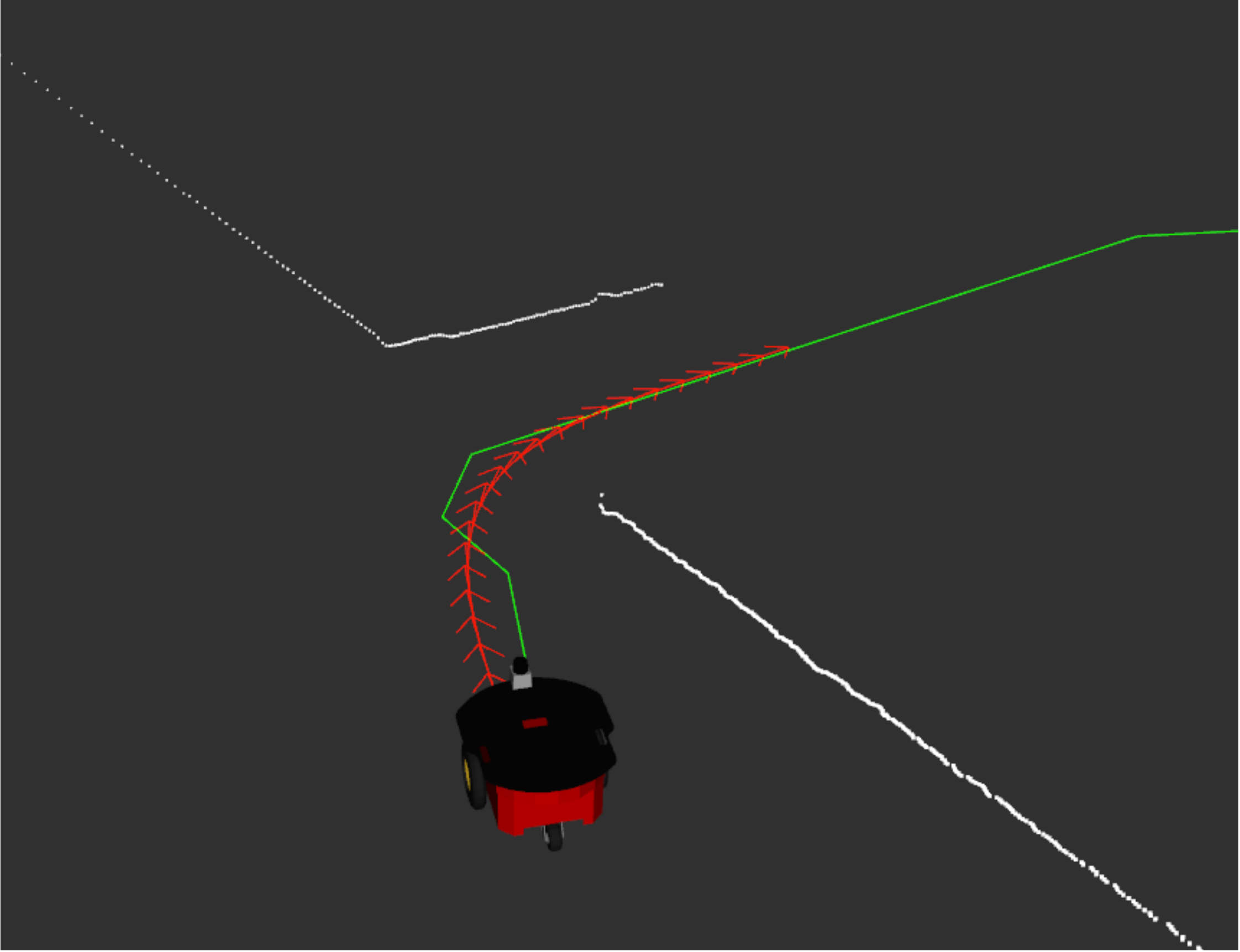}}
	\centering
	\subfigure[]{\includegraphics[scale=0.18]{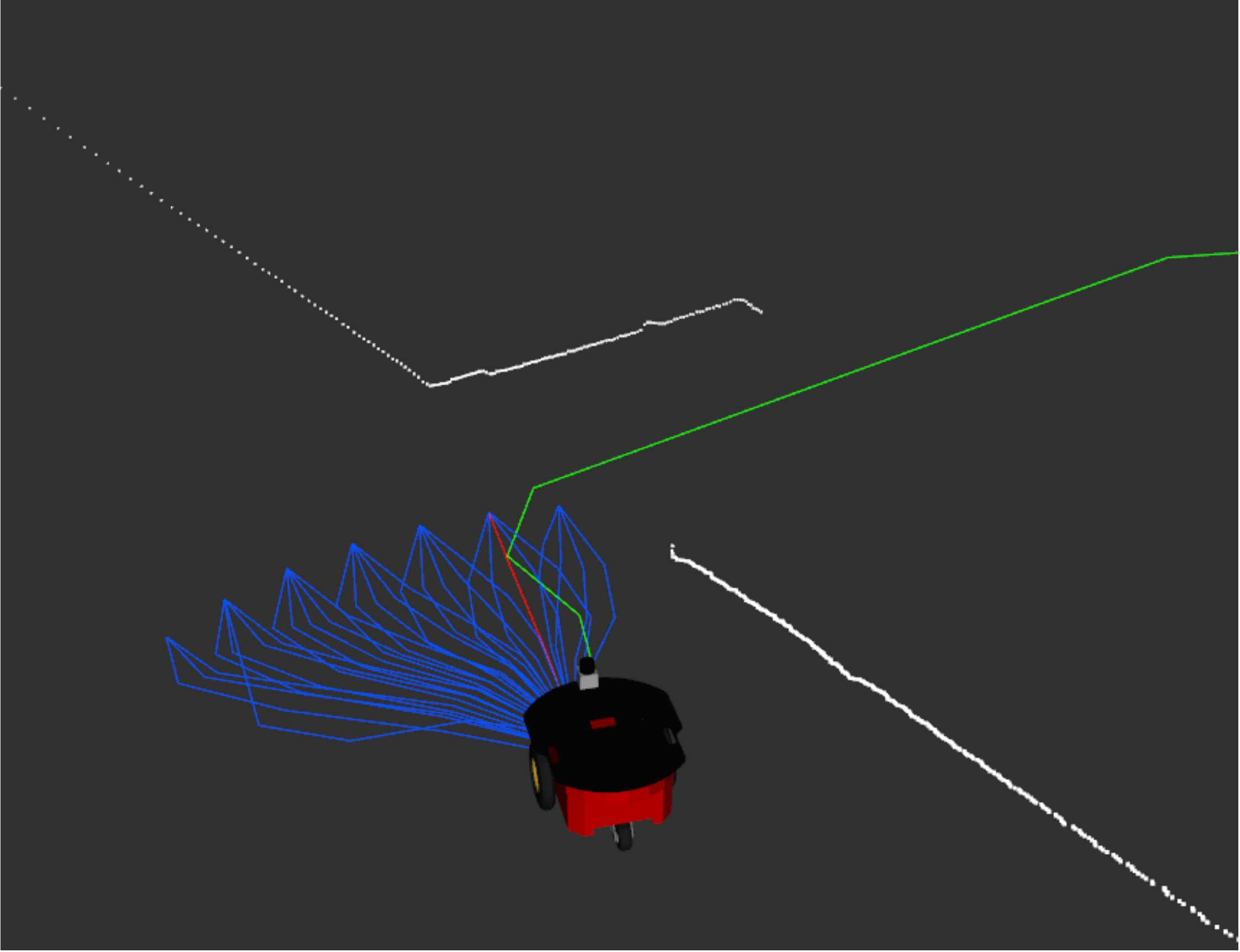}}
	\centering
	\subfigure[]{\includegraphics[scale=0.18]{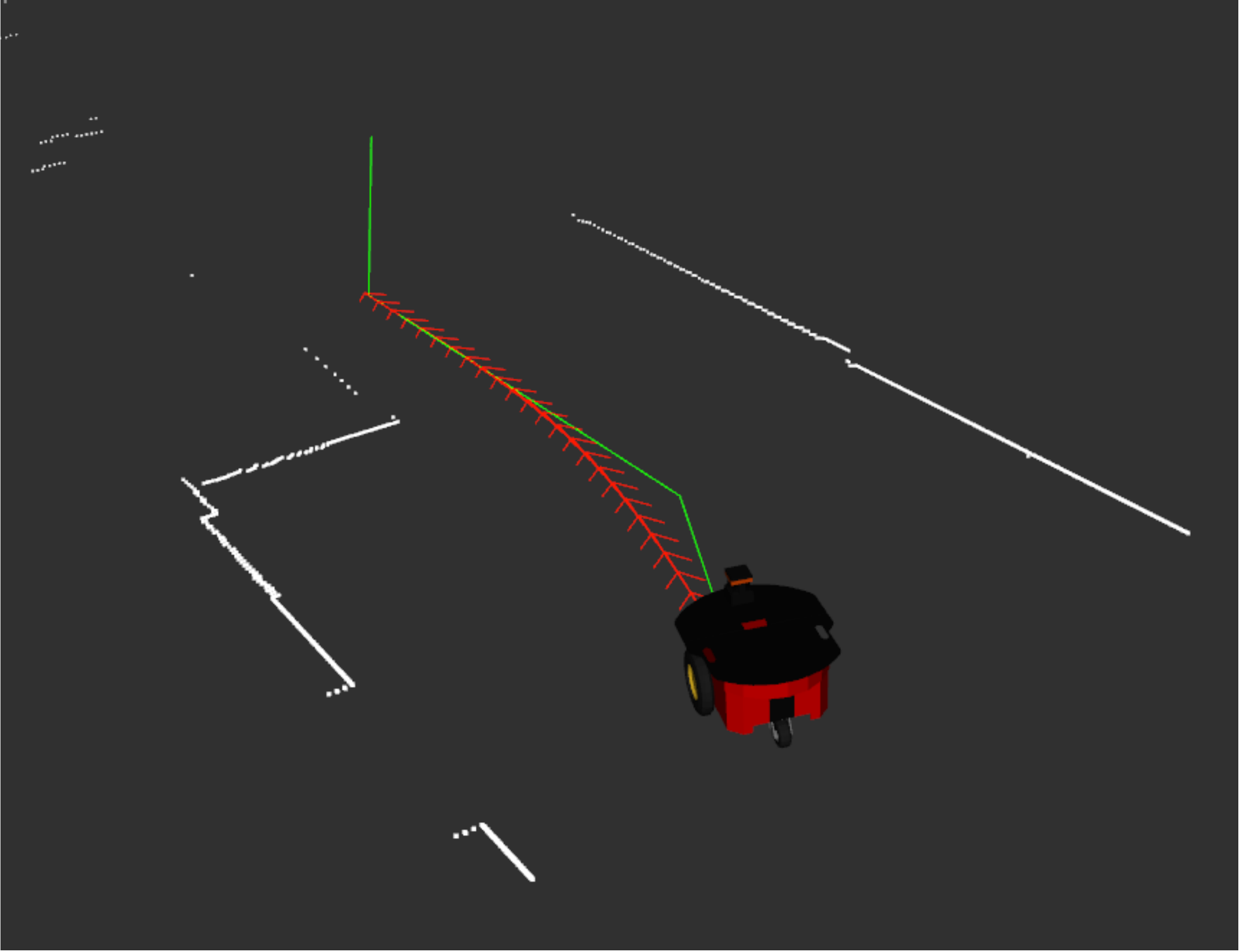}}
	\centering
	\subfigure[]{\includegraphics[scale=0.18]{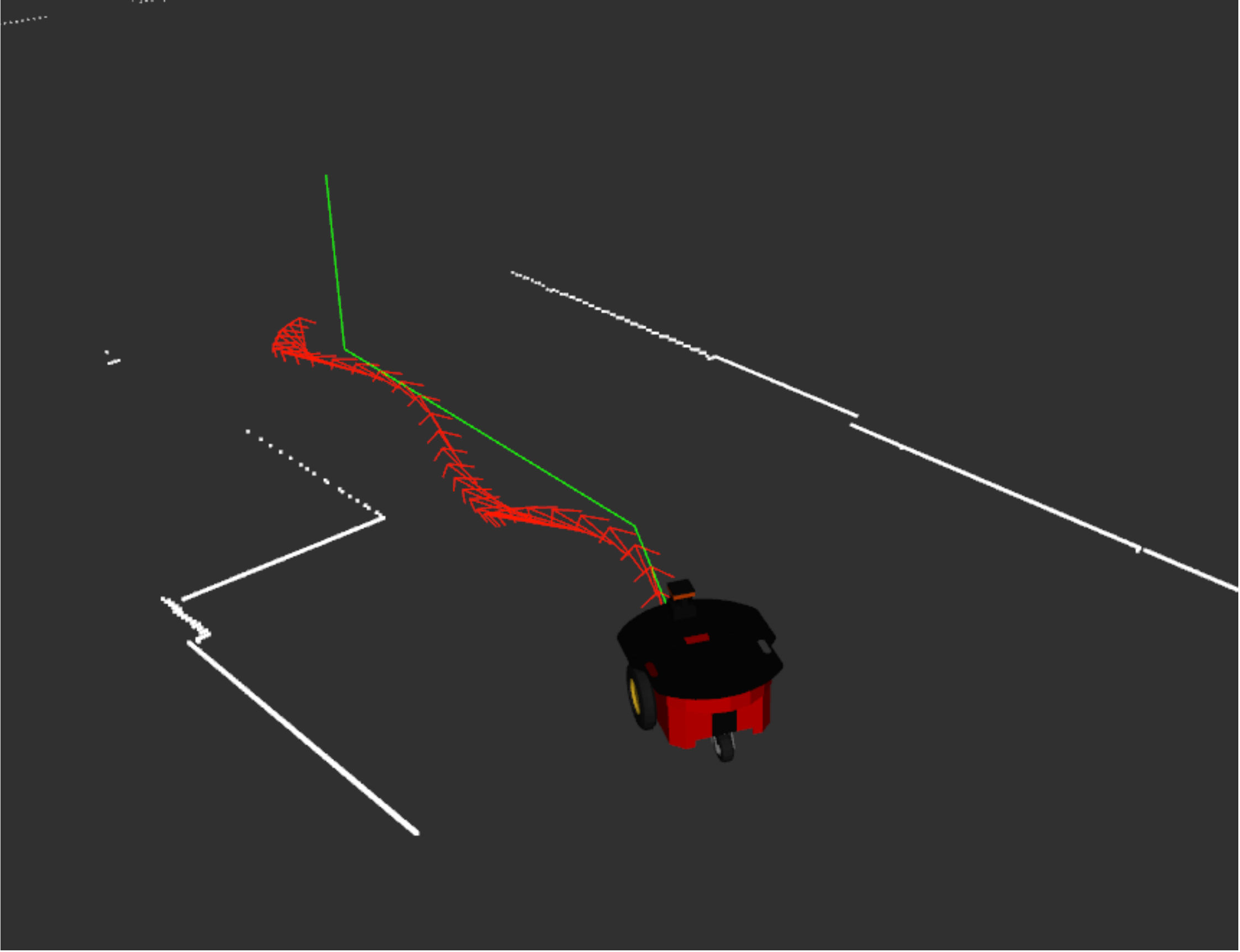}}
	\caption{(a) Experimental scenario II. (b) Planning result of the proposed local planner. (c) Planning result of QBC. (d) Planning result of the proposed local planner. (e) Planning result of TEB.}
	\label{fig:environment}
\end{figure}

\subsection{Comparison on Local Planning}
To validate the flexibility, smoothness, and efficiency of the proposed local planner, Scenario II is designed as illustrated in Fig. \ref{fig:environment}(a). In this scenario, the robot is required to move along a long corridor and make a sharp right turn to pass through the door. In addition, there are two boxes placed inside the door as obstacles. Such a scenario requires local planners to provide smooth and flexible motions to guide the robot to the goal.

In the test of Scenario II, the proposed local planner exhibits excellent motion efficiency and flexibility. In particular, when the robot approaches the door and needs to make a sharp right turn, the proposed local planner guides the robot through the door smoothly, as shown in Fig. \ref{fig:environment}(b). In contrast, QBC guides the robot through the door slowly. As illustrated in Fig. \ref{fig:environment}(c), the endpoints of the offline designed B\'{e}zier curves are fixed and cannot be adjusted according to the real-time planning task, limiting the motion flexibility of local planning. We repeat the experiment several times, and the experimental results are presented in Table \ref{tab:localplanner}. QBC takes approximately $ 25.76 $ $ \mathrm{s} $ to guide the robot to the goal, while the proposed local planner costs only $ 19.86 $ $ \mathrm{s} $ on average in the same situation. Compared with QBC, the motion efficiency of the proposed local planner is improved by $ 22.87\% $. 

To validate the advantage of the proposed path/velocity decoupled local planner, we compare it with the popular open-source path/velocity coupled local planner TEB \cite{rosmann2012trajectory,rosmann2013efficient}. As detailed in Section IV-A, the proposed local path optimization approach is based on a \emph{purely geometric} formulation consisting of smoothness and safety constraints, while the geometric smoothness of the local path is not incorporated in the formulation of TEB. As a result, the local path obtained by the proposed local planner is obviously smoother than that of TEB, as shown in Fig. \ref{fig:environment}(d) and (e). We also repeat the experiment several times, and the experimental results are presented in Table \ref{tab:localplanner}. TEB takes $ 23.75 $ $ \mathrm{s} $ on average to guide the robot to the goal. Compared with TEB, the motion efficiency of the proposed local planner is improved by $ 16.37\% $.

Based on the above comparative experimental results, it is concluded that the proposed local planner achieves superior performance in smoothness, motion efficiency, and flexibility.

\begin{figure}[t]
	\centering
	\includegraphics[scale=0.2]{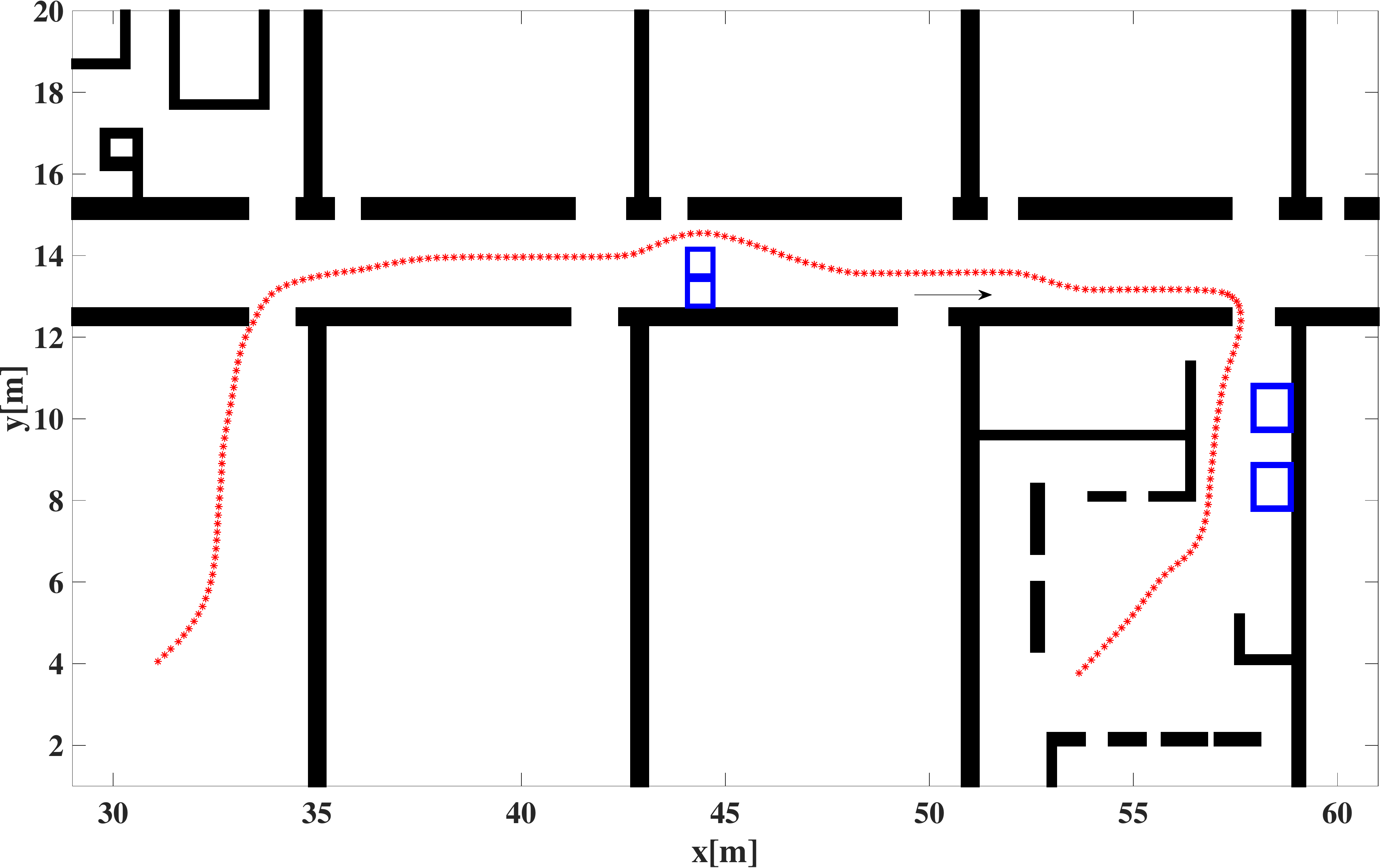}
	\caption{Route of the navigation experiment.}
	\label{fig:navigation}
\end{figure}

\begin{figure}[t]
	\centering
	\subfigure[]{\includegraphics[scale=0.29]{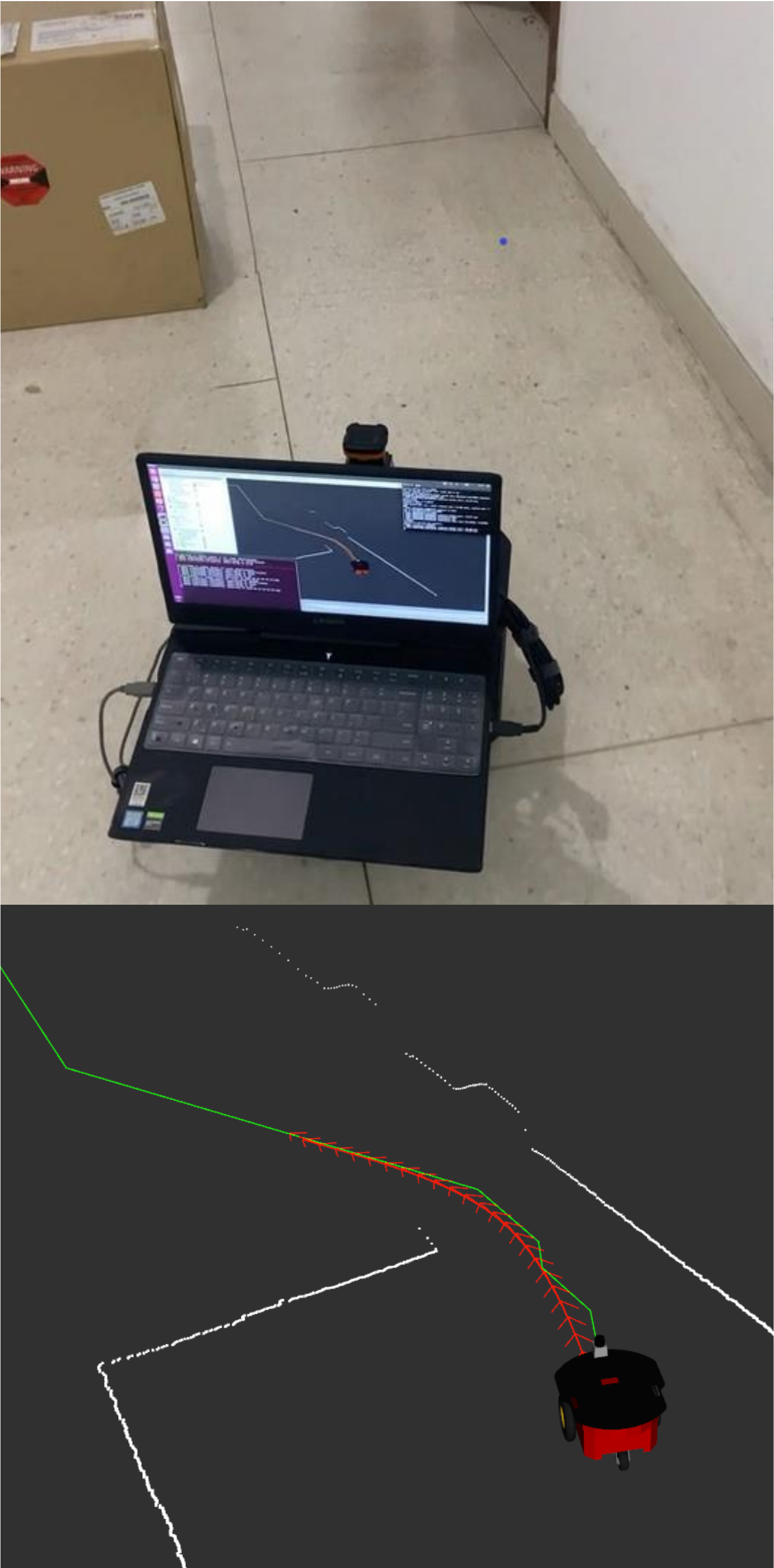}}
	\centering
	\subfigure[]{\includegraphics[scale=0.29]{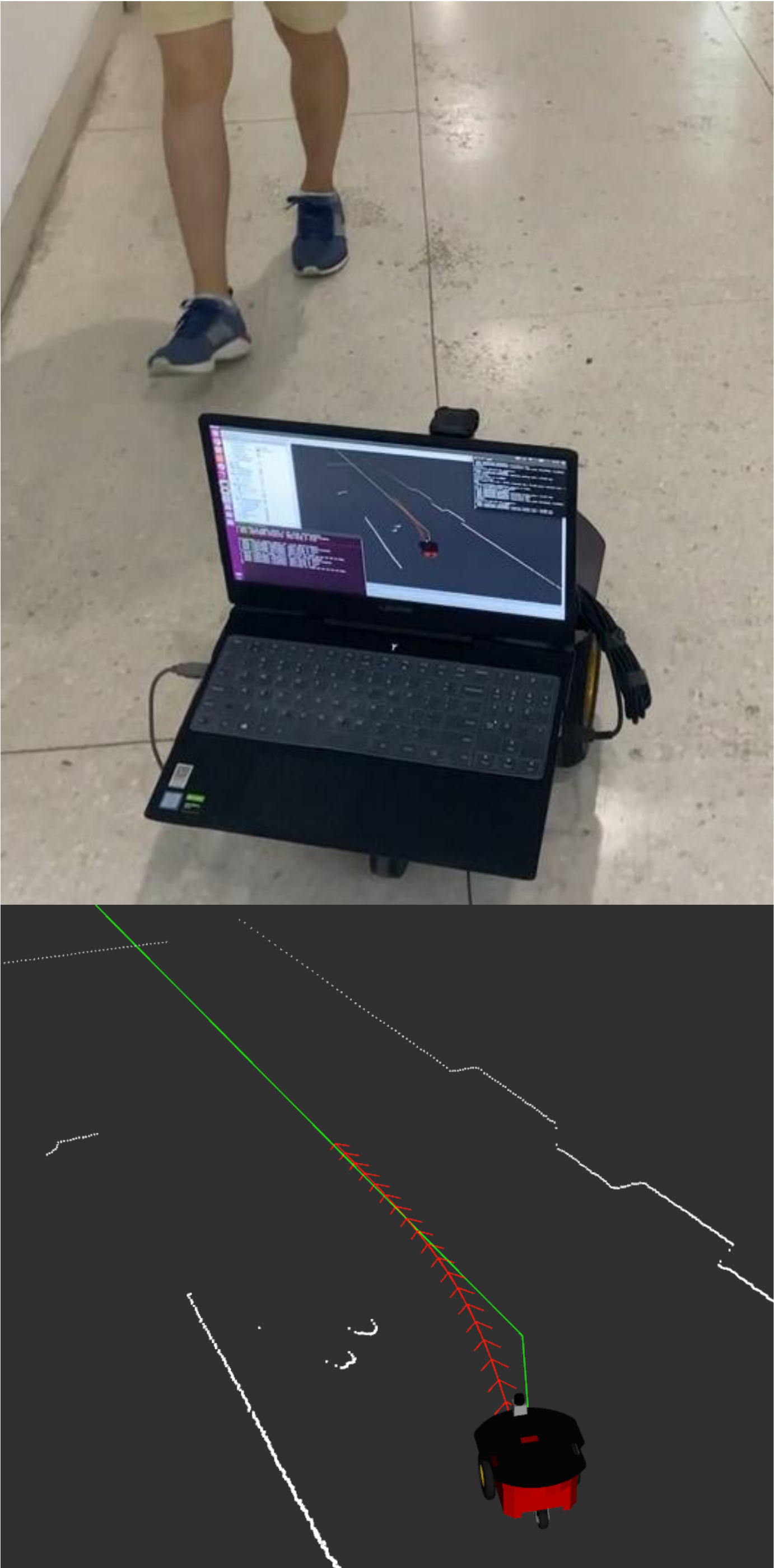}}
	\caption{Experimental scenario III. (a) Passing through a narrow gap. (b) Avoiding an oncoming person.}
	\label{fig:dynamic}
\end{figure}

\subsection{Navigation Experiments}
Finally, we test E\ensuremath{^3}MoP in a $ 92.9 \times 26.5 $ $ \mathrm{m^2} $ laboratory environment. As shown in Fig. \ref{fig:navigation}, the robot is required to move from $ \left( {53.7,3.8} \right) $ to $ \left( {31.0,4.0} \right) $. To make the test more challenging, we place some boxes as static obstacles in the long corridor that the robot passes through. Furthermore, we also design a dynamic pedestrian scene to challenge the safety, flexibility, and robustness of E\ensuremath{^3}MoP. The total travel distance of the navigation experiment is approximately $ 44.2 $ $ \mathrm{m} $, and E\ensuremath{^3}MoP takes $ 62.83 $ $ \mathrm{s} $ to guide the robot to the goal.

\begin{figure}[t]
	\centering
	\includegraphics[scale=0.19]{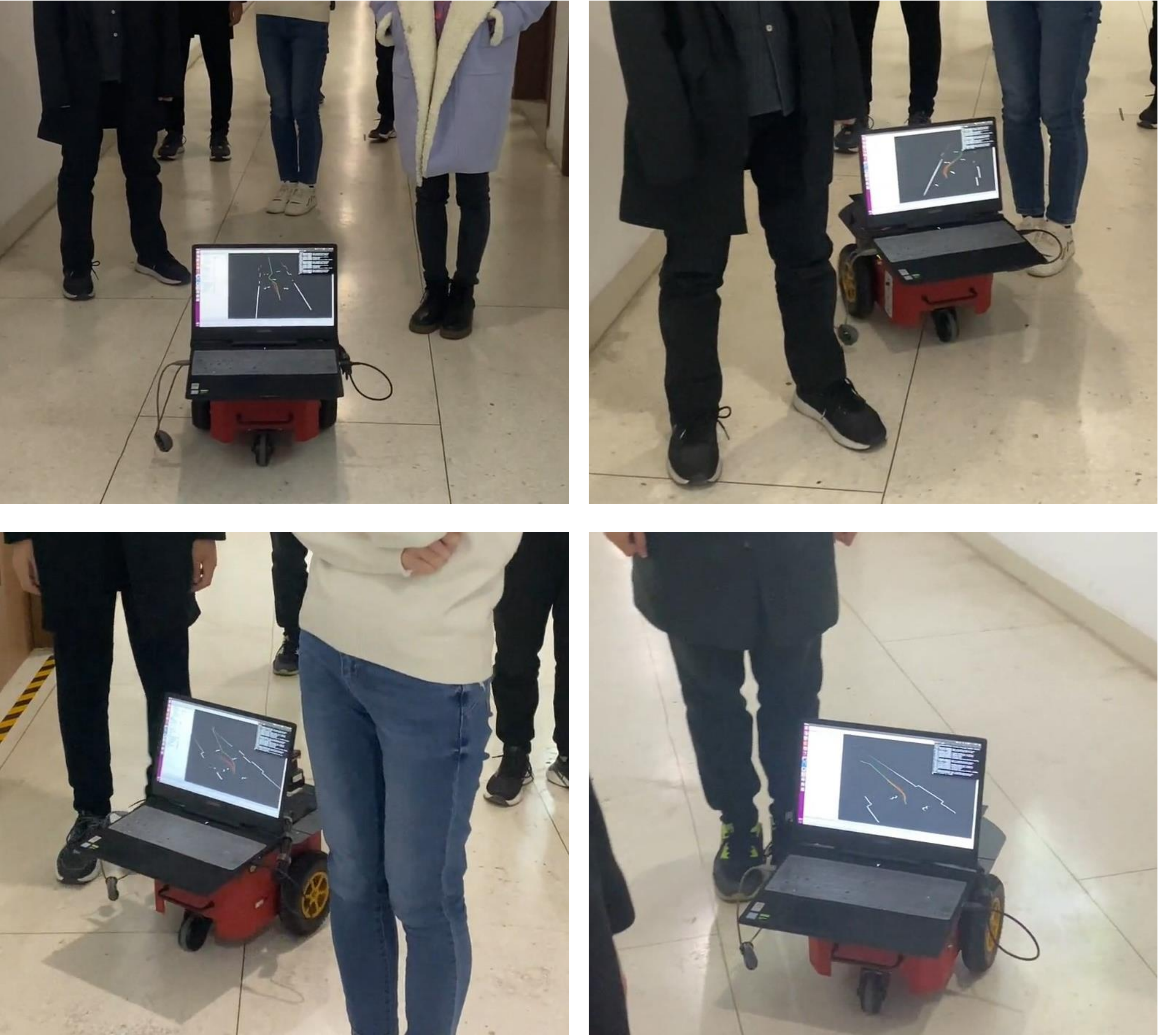}
	\caption{Screenshots of the robot passing through the crowd.}
	\label{fig:crowd}
\end{figure}

Here we summarize several representative experimental results of the navigation experiment to demonstrate the key characteristics of E\ensuremath{^3}MoP.

\subsubsection{Dealing with static obstacles}
Fig. \ref{fig:dynamic}(a) illustrates the testing scenario with static obstacles. The robot avoids some boxes and passes through a narrow gap smoothly, according to the reliable local planning results.

\subsubsection{Dealing with dynamic obstacles}
Fig. \ref{fig:dynamic}(b) shows the testing scenario with dynamic obstacles. The robot implements fast re-planning and avoids an oncoming person successfully, thanks to the efficient local planner.

\subsubsection{Passing through the crowd}
Fig. \ref{fig:crowd} shows the robot smoothly passes through the crowd under the guidance of the local planner. This challenging scenario requires the local planner to provide flexible, safe, and smooth motion commands.

\section{Conclusion}
In this paper, a three-layer motion planning framework called E\ensuremath{^3}MoP is proposed to address the motion planning problem of mobile robots in complex environments. A novel heuristic-guided pruning strategy of motion primitives is proposed to improve the computational efficiency of graph search. And a soft-constrained local path optimization approach combined with time-optimal velocity planning is presented to generate safe, smooth, and efficient motion commands according to real-time sensor data. Furthermore, the sparse-banded system structure of the underlying path optimization formulation is fully exploited to efficiently solve the problem. Extensive simulations and experiments are presented and discussed to validate that E\ensuremath{^3}MoP has superior performance in terms of safety, smoothness, flexibility, and efficiency.
	
The objective function of the local path optimization that results from convex and concave terms is non-convex, thus the local planner may get stuck in local optima and is unable to transit across obstacles. In the future, we plan to extend the proposed local path optimization approach with the theory of homology classes \cite{bhattacharya2012topological} to maintain several homotopically distinct local paths and seek global optima.

\section*{Acknowledgment}
The data for the Intel Research Lab is available from the Radish data set repository \cite{howard2003robotics}. The authors gratefully thank Dirk H{\"a}hnel for providing this data set. 

\bibliographystyle{IEEEtran}
\bibliography{ref}

\begin{IEEEbiography}[{\includegraphics[width=1in,height=1.25in,clip,keepaspectratio]{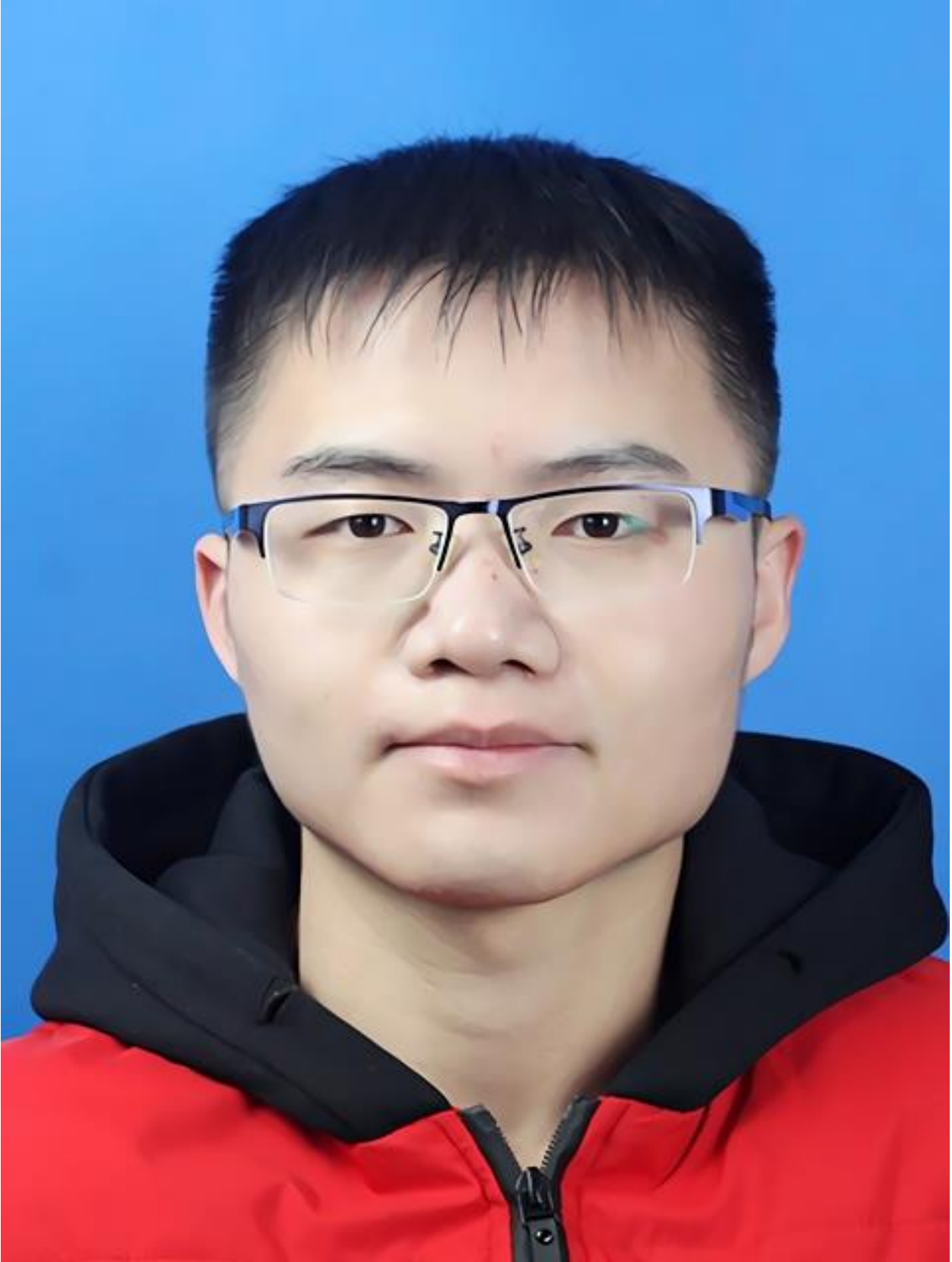}}]{Jian Wen} received the B.S. degree in automation from Nankai University, Tianjin, China, in 2017. He is currently working toward the Ph.D. degree in control science and engineering with the Institute of Robotics and Automatic Information System, Nankai University.
	
His research interests include mapping and motion planning for mobile robots.
\end{IEEEbiography}

\begin{IEEEbiography}[{\includegraphics[width=1in,height=1.25in,clip,keepaspectratio]{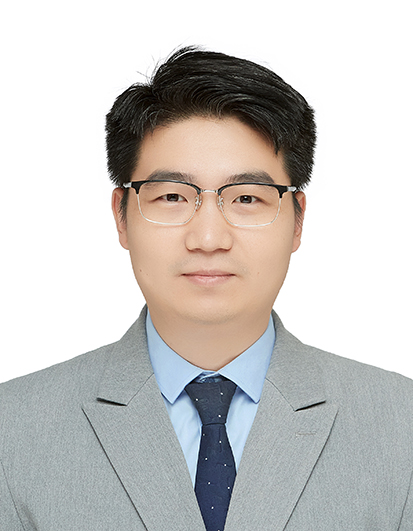}}]{Xuebo Zhang} (M'12$-$SM'17) received the B.Eng. degree in automation from Tianjin University, Tianjin, China, in 2006, and the Ph.D. degree in control theory and control engineering from Nankai University, Tianjin, China, in 2011.
	
From 2014 to 2015, he was a Visiting Scholar with the Department of Electrical and Computer Engineering, University of Windsor, Windsor, ON, Canada. He was a Visiting Scholar with the Department of Mechanical and Biomedical Engineering, City University of Hong Kong, Hong Kong, in 2017. He is currently a Professor with the Institute of Robotics and Automatic Information System, Nankai University, and Tianjin Key Laboratory of Intelligent Robotics, Nankai University. His research interests include planning and control of autonomous robotics and mechatronic system with focus on time-optimal planning and visual servo control; intelligent perception including robot vision, visual sensor networks, SLAM; reinforcement learning and game theory.
	
Dr. Zhang is a Technical Editor of \emph{IEEE/ASME Transactions on Mechatronics} and an Associate Editor of \emph{ASME Journal of Dynamic Systems, Measurement, and Control}.
\end{IEEEbiography}

\begin{IEEEbiography}[{\includegraphics[width=1in,height=1.25in,clip,keepaspectratio]{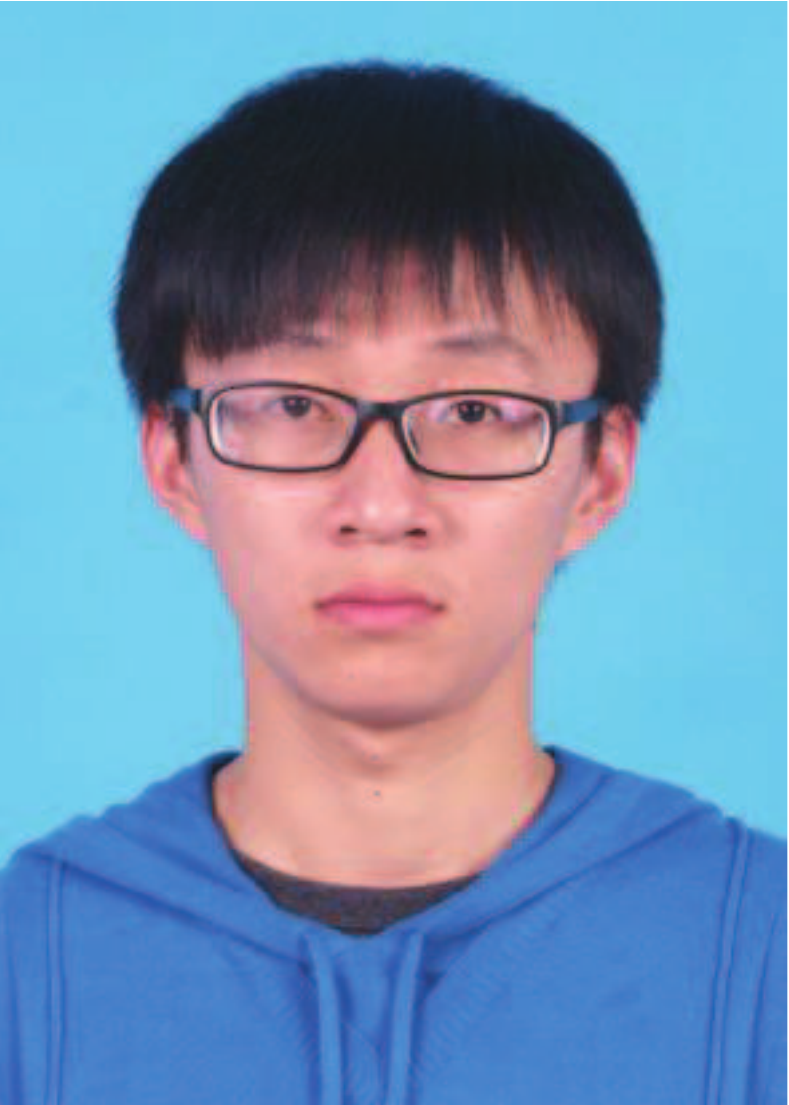}}]{Haiming Gao} received the B.S. degree in mechanical engineering and automation from Hohai University, Nanjing, China, in 2015, and the Ph.D. degree in control science and engineering from Nankai University, Tianjin, China, in 2020. From 2020 to 2021, he worked as a Senior Algorithm Engineer in IAS BU (Huawei Intelligent Automotive Solution), with research interests include freespace extraction, APA (Auto Parking Asist), and AVP (Automated Valet Parking). He is currently a Research Fellow with Zhejiang Lab.
	
His research interests include localization and perception for intelligent vehicles and mobile robots.
\end{IEEEbiography}

\begin{IEEEbiography}[{\includegraphics[width=1in,height=1.25in,clip,keepaspectratio]{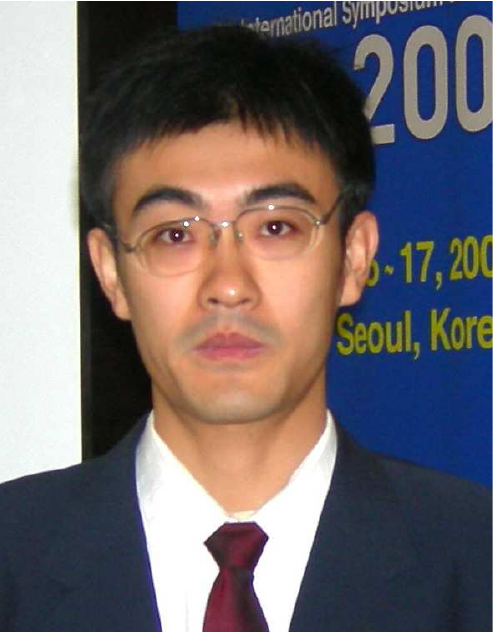}}]{Jing Yuan} (M'12) received the B.S. degree in automatic control, and the Ph.D. degree in control theory and control engineering from Nankai University, Tianjin, China, in 2002 and 2007, respectively.
	
Since 2007, he has been with the College of Computer and Control Engineering, Nankai University, where he is currently a Professor. His current research interests include robotic control, motion planning, and SLAM.
\end{IEEEbiography}

\begin{IEEEbiography}[{\includegraphics[width=1in,height=1.25in,clip,keepaspectratio]{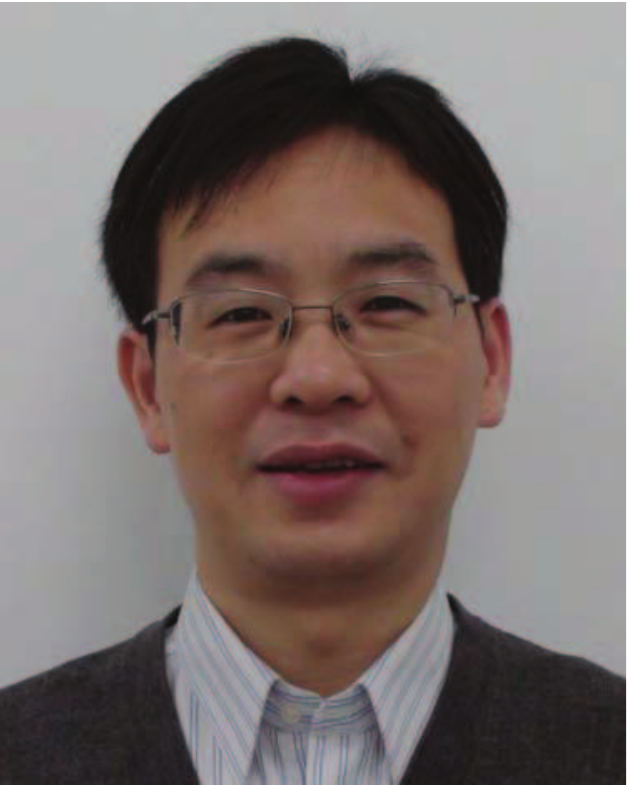}}]{Yongchun Fang} (S'00$-$M'02$-$SM'08) received the B.S. degree in electrical engineering and the M.S. degree in control theory and applications from Zhejiang University, Hangzhou, China, in 1996 and 1999, respectively, and the Ph.D. degree in electrical engineering from Clemson University, Clemson, SC, in 2002.
	
From 2002 to 2003, he was a Post-Doctoral Fellow with the Sibley School of Mechanical and Aerospace Engineering, Cornell University, Ithaca, NY, USA. He is currently a Professor with the Institute of Robotics and Automatic Information System, Nankai University, Tianjin, China. His research interests include visual servoing, AFM-based nano-systems, and control of underactuated systems including overhead cranes.
\end{IEEEbiography}

\end{document}